\documentclass[sigconf, nonacm]{acmart}

\newcommand\vldbdoi{XX.XX/XXX.XX}
\newcommand\vldbpages{XXX-XXX}
\newcommand\vldbvolume{14}
\newcommand\vldbissue{1}
\newcommand\vldbyear{2020}
\newcommand\vldbauthors{\authors}
\newcommand\vldbtitle{\shorttitle} 
\newcommand\vldbavailabilityurl{URL_TO_YOUR_ARTIFACTS}
\newcommand\vldbpagestyle{plain} 

\usepackage{graphicx}
\usepackage{amsmath}
\usepackage[linesnumbered,ruled,vlined]{algorithm2e}
\usepackage{subcaption}
\usepackage{booktabs}  
\usepackage{xcolor}    

\begin{document}
\title{Krul: Efficient State Restoration for Multi-turn Conversations with Dynamic Cross-layer KV Sharing}

\author{Junyi Wen}
\affiliation{%
  \institution{Sun Yat-sen University}
  \streetaddress{P.O. Box 1212}
  \city{Zhuhai}
  \state{China}
  \postcode{43017-6221}
}
\email{wenjy23@mail2.sysu.edu.cn}

\author{Junyuan Liang}
\orcid{0000-0002-1825-0097}
\affiliation{%
  \institution{Sun Yat-sen University}
  \streetaddress{1 Th{\o}rv{\"a}ld Circle}
  \city{Guangzhou}
  \country{China}
}
\email{liangjy53@mail2.sysu.edu.cn}

\author{Zicong Hong}
\orcid{0000-0001-5109-3700}
\affiliation{%
  \institution{Hong Kong University of Science and Technology}
  \city{Hong Kong}
  \country{China}
}
\email{congcong@ust.hk}

\author{Wuhui Chen}
\affiliation{%
  \institution{Sun Yat-sen University}
  \city{Zhuhai}
  \country{China}\\
  \institution{Peng Cheng Laboratory}
  \city{Shenzhen}
  \country{China}
}
\email{chenwuh@mail.sysu.edu.cn}

\author{Ting Cai}
\orcid{0000-0002-1825-0097}
\affiliation{%
  \institution{Hubei University of Technology}
  \streetaddress{1 Th{\o}rv{\"a}ld Circle}
  \city{Wuhan}
  \country{China}
}
\email{caiting@hbut.edu.cn}

\author{Zibin Zheng}
\affiliation{%
  \institution{Sun Yat-sen University}
  \city{Zhuhai}
  \country{China}
}
\email{zhzibin@mail.sysu.edu.cn}

\begin{abstract}

Efficient state restoration in multi-turn conversations with large language models (LLMs) remains a critical challenge, primarily due to the overhead of recomputing or loading full key-value (KV) caches for all historical tokens. To address this, existing approaches compress KV caches across adjacent layers with highly similar attention patterns. However, these methods often apply a fixed compression scheme across all conversations, selecting the same layer pairs for compression without considering conversation-specific attention dynamics. This static strategy overlooks variability in attention pattern similarity across different conversations, which can lead to noticeable accuracy degradation.

We present Krul, a multi-turn LLM inference system that enables accurate and efficient KV cache restoration. Krul dynamically selects compression strategies based on attention similarity across layer pairs and uses a recomputation-loading pipeline to restore the KV cache. It introduces three key innovations: 1) a \textit{preemptive compression strategy selector} to preserve critical context for future conversation turns and selects a customized strategy for the conversation; 2) a \textit{token-wise heterogeneous attention similarity estimator} to mitigate the attention similarity computation and storage overhead during model generation; 3) a \textit{bubble-free restoration scheduler} to reduce potential bubbles brought by the imbalance of recomputing and loading stream due to compressed KV caches. Empirical evaluations on real-world tasks demonstrate that Krul achieves a 1.5$\times$ $\sim$ 2.68$\times$ reduction in time-to-first-token (TTFT) and a 1.33$\times$ $\sim$ 2.35$\times$ reduction in KV cache storage compared to state-of-the-art methods without compromising generation quality.
\end{abstract}

\maketitle

\pagestyle{\vldbpagestyle}
\begingroup\small\noindent\raggedright\textbf{PVLDB Reference Format:}\\
\vldbauthors. \vldbtitle. PVLDB, \vldbvolume(\vldbissue): \vldbpages, \vldbyear.\\
\href{https://doi.org/\vldbdoi}{doi:\vldbdoi}
\endgroup
\begingroup
\renewcommand\thefootnote{}\footnote{\noindent
This work is licensed under the Creative Commons BY-NC-ND 4.0 International License. Visit \url{https://creativecommons.org/licenses/by-nc-nd/4.0/} to view a copy of this license. For any use beyond those covered by this license, obtain permission by emailing \href{mailto:info@vldb.org}{info@vldb.org}. Copyright is held by the owner/author(s). Publication rights licensed to the VLDB Endowment. \\
\raggedright Proceedings of the VLDB Endowment, Vol. \vldbvolume, No. \vldbissue\ %
ISSN 2150-8097. \\
\href{https://doi.org/\vldbdoi}{doi:\vldbdoi} \\
}\addtocounter{footnote}{-1}\endgroup

\ifdefempty{\vldbavailabilityurl}{}{
\vspace{.3cm}
\begingroup\small\noindent\raggedright\textbf{PVLDB Artifact Availability:}\\
The source code, data, and/or other artifacts have been made available at \url{https://github.com/xxxx/xxxx}.
\endgroup
}

\section{Introduction}

The ability to engage in multi-turn conversations—characterized by multiple rounds of user queries and model responses—is a core strength of large language models (LLMs)~\cite{kalyan2024survey, chang2024survey}. Such interactions enable models to better capture user intent and provide more contextually appropriate responses~\cite{feng2022mmdialog, wang2021naturalconv}. An analysis of ShareGPT~\cite{chen2024sharegpt4v}, a real-world dataset collected from ChatGPT, reveals that over 70\% of conversations involve multiple turns, highlighting the prevalence and importance of multi-turn conversations.

Despite the importance, efficiently serving multi-turn conversations remains a significant challenge~\cite{ye2025flashinfer, kwon2023efficient}. When generating the response during a turn of conversation, the LLM inference system caches intermediate states, referred to as \textit{key-value (KV) caches}~\cite{zhang2024kv, adnan2024keyformer, hooper2024kvquant}, to avoid the redundant computations of historical tokens. However, once a conversation ends and becomes inactive, these KV caches are typically evicted to free GPU memory~\cite{yogatama2024scaling}. When the user resumes the conversation with a new input, the system must prefill the latest prompt along with the full conversation history and recompute the KV cache of all historical tokens to restore the conversation states. This results in extensive redundant computation and significantly increases both time-to-first-token (TTFT) and overall response latency, undermining the responsiveness of real-time applications~\cite{fan2024combining, fu2024encchain, zhang2024applications}.

\begin{figure}[t]
    \vspace{-8pt}
    \centering
\setlength\abovecaptionskip{0.02\baselineskip}
\setlength\belowcaptionskip{-0.5\baselineskip}
	\includegraphics[width=\linewidth]{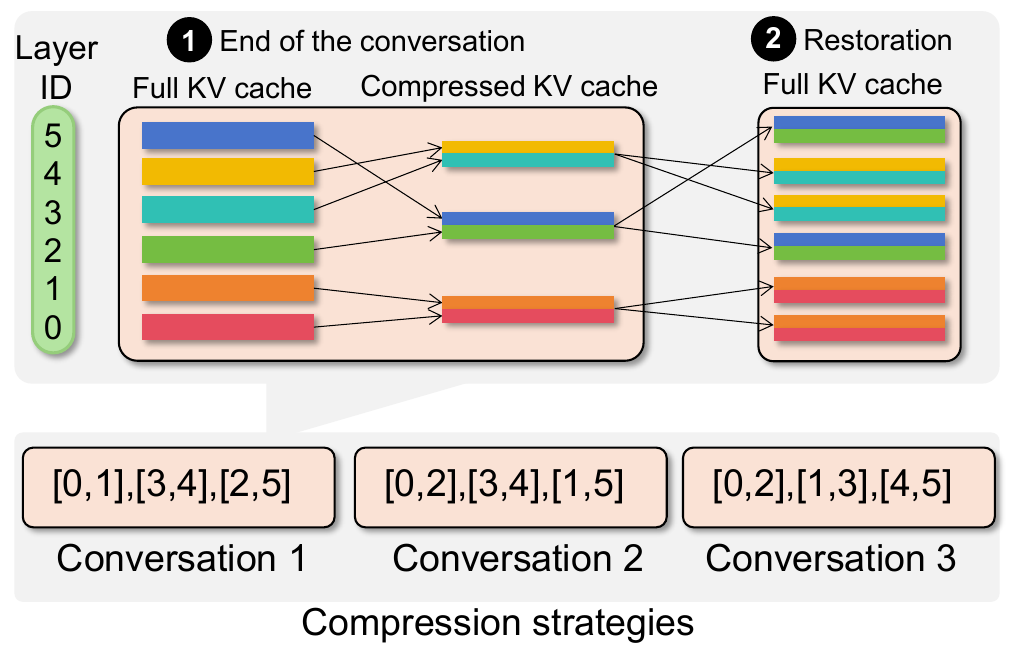}
	\caption{Overview of Krul's state restoration. Conversations have customized KV caches compression strategies. [0,1] means compressing KV cache across layer 0 and 1.}
    \label{Krul_strawman_overview}
    \vspace{-8pt}
\end{figure}

\textbf{Limitations of Existing Methods.} In multi-turn conversations inference, offloading KV caches to CPU memory for inactive conversations demonstrates higher efficiency compared to recomputation. Currently, two predominant loading approaches exist: \textit{full KV cache loading} \cite{gao2024cost, gao2025fast, kwon2023efficient, yu2025stateful, jin2024compute, yin2024llm} and \textit{compressed KV cache loading}~\cite{liu2024minicache, yang2024kvsharer, mu2024cross}. The full KV cache loading approach stores the complete set of KV caches for inactive conversations to enable rapid restoration. However, the memory space of the KV cache is still costly. For example, Qwen1.5-7B takes an average of 1GB of KV cache space for each conversation turn with 2000 tokens. As a result, around 50 multi-turn conversations will soon take up the memory space of the CPU.  To reduce the KV cache storage, existing works observe that the attention similarity between deep adjacent layer pairs is very high. Therefore, they use compressed KV cache loading by compressing the KV caches across these layer pairs. However,  the compressed KV cache loading approach often leads to noticeable degradation in generation quality for certain conversation contexts. That is because the compression strategy is fixed, typically applying uniform compression across every two adjacent deep layers, failing to account for the heterogeneous distribution of information across different conversations. 


\textbf{Compression strategies should be adaptive.}  We find that the attention similarities vary across different conversations. Our empirical study, involving three distinct randomized KV cache compression strategies, reveals that the optimal strategy is context-dependent - no single method achieves universal superiority across all conversations (detailed in Section~\ref{strawman}). We explain the example in Figure \ref{Krul_strawman_overview}: strategy 1 (layers [0, 1], [2, 5], [3, 4]) works well for conversation 1 but causes significant accuracy loss in conversation 2, where strategy 2 (layers [0, 2], [1, 5], [3, 4]) performs better due to higher attention similarity of the layer pairs in the strategy. Therefore,  an adaptive approach—dynamically adjusting the KV cache compression strategy based on conversational characteristics—could potentially balance the trade-offs between efficiency and quality. Based on this observation, we introduce a novel inference system that dynamically selects a conversation-adaptive compression strategy based on attention similarity. However, this system still faces significant challenges.

\textbf{(1) Compression without Knowledge of Future User Input.} In multi-turn conversations, future user inputs are inherently unpredictable, making it difficult to anticipate which portions of the historical context will be referenced during generation. The attention weight of the same token varies with different user inputs. As a result, aggressive or ill-targeted compression can degrade critical contextual information and compromise generation quality. \textbf{(2) Overhead Introduced by the Compression Process.} Selecting adaptive compression strategies requires attention weights storage, similarity computation, incurring non-trivial overhead during generation that compounds with the $O(n^2)$ scaling of attention weights. Moreover, directly saving all attention weights in the GPU can easily trigger an Out-of-Memory error, which requires 30GB of GPU memory for a prompt with a length of 4000 tokens in a 7B model. \textbf{(3) Efficient Restoration Using Compressed KV Caches.} Standard state restoration typically relies on a recomputation-loading
approach~\cite{yu2025stateful, jin2024compute, yin2024llm}, which assumes a fixed pattern of token recomputation and KV cache loading. However, this pipeline can suffer from imbalanced computation and I/O streams due to the compressed KV caches.


To address these challenges, we propose Krul, a cost-efficient inference system that leverages conversation-adaptive KV cache compression to enable fast and high-quality multi-turn state restoration for LLMs. It introduces three core components that collectively address the three challenges. (1) Preemptive compression strategy selector. For the first time, we observe attention patterns of some model layers vary with different user inputs, while others do not. To preserve critical context while adapting compression to the characteristics of each conversation, Krul proposes a preemptive attention pattern analysis algorithm to identify input-sensitive model layers that are unsuitable for compression. After extracting these layers,  Krul selects customized KV cache compression strategies tailored to each historical conversation. (2) Token-wise heterogeneous attention similarity estimator. The attention weights of the prefilling phase are large—covering all input tokens—and consume significant GPU memory, but are fast to compute similarity. In contrast, attention weights of the decoding phase are small—only focused on the latest token—but require frequent similarity computation, as decoding typically generates an average of ~6.56$\times$ more tokens than the prefilling phase. To balance memory and time efficiency, Krul applies a token-wise partition to the attention weights. Krul offloads attention weights of the prefilling phase to the CPU for asynchronous similarity computation, while retaining attention weights of the decoding phase on the GPU for token-wise computation. The final attention similarity is obtained by aggregating both the results, achieving accurate estimation with minimal GPU overhead. (3) Bubble-free restoration scheduler. To address imbalances between the recomputation and loading stages in the presence of compressed KV caches, Krul introduces a pipeline orchestration mechanism to dynamically schedule recomputation and loading tasks across model layers, ensuring efficient overlap and end-to-end performance.

This work makes the following contributions:

\begin{itemize}
\item We identify a significantly broader KV cache compression strategy selection space enabled by attention similarity patterns in LLMs, and demonstrate that fixed compression strategies are suboptimal for multi-turn conversations.
\item We propose Krul, a conversation-adaptive inference system that incorporates three novel components—preemptive compression strategy selector, token-wise heterogeneous attention similarity estimator, and bubble-free restoration scheduler—to enable fast and memory-efficient state restoration.
\item We implement Krul using Python and PyTorch, and evaluate it on real-world multi-turn conversation benchmarks. Krul achieves a 1.28$\times$ $\sim$ 2.68$\times$ reduction in time-to-first-token (TTFT) and a 1.33$\times$ $\sim$ 2.35$\times$ reduction in KV cache storage, compared to state-of-the-art baselines, without compromising generation quality.
\end{itemize}

\section{Background and Motivation}

This section first introduces the fundamentals of generative LLM inference and explores the inefficiency of multi-turn conversation state restoration, and then discusses the design opportunities for dealing with these inefficiencies and the challenges faced in designing such a system.

\subsection{Generative LLM Inference Basics}

\textbf{Transformer Architecture.} Many LLMs adopt the transformer architecture~\cite{vaswani2017attention, han2021transformer} as their basic building block. The widely used LLMs like DeepSeek~\cite{liu2024deepseek}, GPT~\cite{kalyan2024survey}, Qwen~\cite{bai2023qwen}, and LLaMA~\cite{touvron2023llama} are built upon the transformer architecture. The user's prompt is tokenized~\cite{choo2023study, rust2020good} to a sequence of input tokens. Then the model predicts the probability of subsequent tokens and generates the response to the user~\cite{yenduri2024gpt, yenduri2023generative}. A model consists of multiple transformer layers. A transformer layer comprises two major components: the \textit{attention module}~\cite{gheini2021cross} and the \textit{feed-forward network (FFN)}~\cite{touvron2022resmlp, moller2024efficient}.


\begin{figure}[t]
\centering
    \begin{subfigure}{0.49\linewidth}
            \centering
            \includegraphics[width=1.0\linewidth]{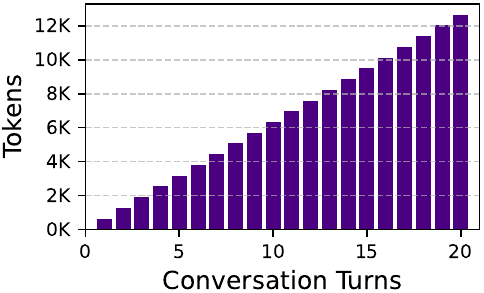}
            \caption{The total prompt length with the increase of conversation turns in ShareGPT.}
            \label{token_turn}
        \end{subfigure}
        \begin{subfigure}{0.49\linewidth}
        \centering
        \includegraphics[width=1.0\linewidth]{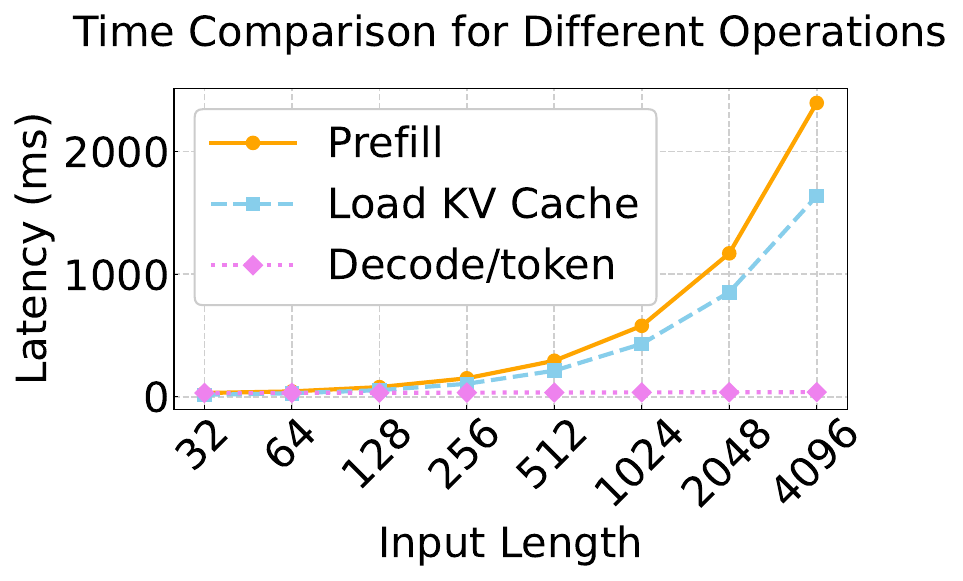}
        \caption{Comparison of prefilling, KV caches loading and decoding latency.}
        \label{prefill_pingjing}
    \end{subfigure}
    
    \caption{The analysis of ShareGPT.}
    \label{prefilling_problem}
\end{figure}

For the input prompt, firstly the model tokenizes it to an input token list $X=[x_1, x_2, \cdots, x_s]$. For hidden state input $X_l$ in each layer $l$, each layer applies a series of projections on each token in $X_l$ using the weights $W^Q_l, W^K_l, W^V_l$. This generates the \textit{queries}, \textit{keys}, and \textit{values} in each layer, referred to as $Q_l$, $K_l$ and $V_l$ respectively:
$$Q_l=W^Q_lX_l, K_l=W^K_lX_l, V_l=W^V_lX_l$$
Next, the model computes the relevance of $x_i$ with all the tokens in front of $x_i$ for each token in $X$ via $Q$ of $x_i$ and $K,V$ of other tokens, referred to as \textit{attention weights} respectively:
$$Attention(Q,K,V)=softmax(\frac{QK^T}{\sqrt{d_K}})V$$
where $d_K$ is the dimension of the key vector $K$. Finally, the result is input to the FFN layer and then passed to the next transformer layer. After the input has been processed through all $N$ transformer layers, the model outputs a probability list that figures out the most probable output tokens.

\textbf{KV caches:} Since all $K$ and $V$ tensors of preceding tokens are necessary for a new-generated token to compute the self-attention, these $K$ and $V$ tensors are cached in GPUs to avoid repeated computation, referred to as the KV caches.


\subsection{Autoregressive Generation and Multi-turn Conversation Inference}

The transformer-based generation consists of two phases: \textit{the prefilling phase} and \textit{the decoding phase}.

\begin{figure}[t]
\centering
    \begin{subfigure}{0.49\linewidth}
        \centering
        \includegraphics[width=1.0\linewidth]{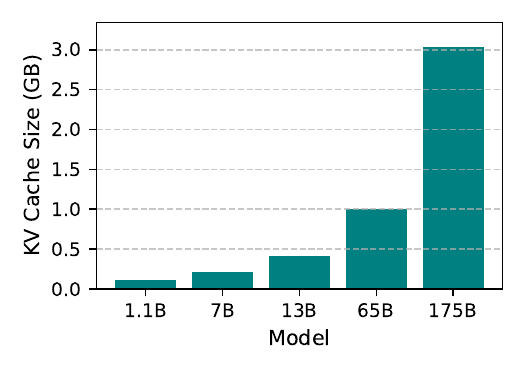}
        \caption{The KV caches storage of each conversation turn varying with different sizes of models.}
        \label{kv_storage}
    \end{subfigure}
    \hfill
    \begin{subfigure}{0.49\linewidth}
        \centering
        \includegraphics[width=1.0\linewidth]{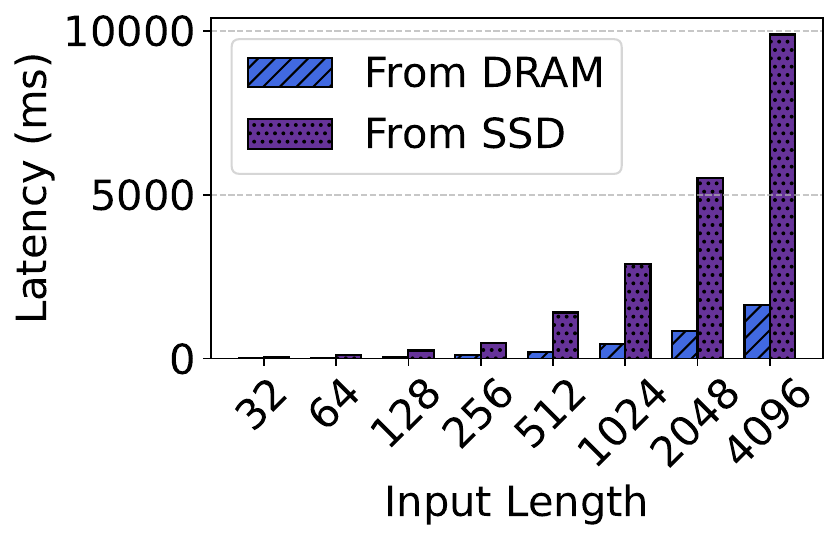}
        \caption{Comparison of the KV caches loading latency from the DRAM and the SSD.}
        \label{KV_load}
    \end{subfigure}
    \caption{The storage and loading overhead of KV caches.}
    \label{prefilling_problem}
    \vspace{-8pt}
\end{figure}

\begin{figure*}[t]
    \centering
    \vspace{-8pt}
    \begin{minipage}{0.25\linewidth}  
        \centering
        \includegraphics[width=\linewidth]{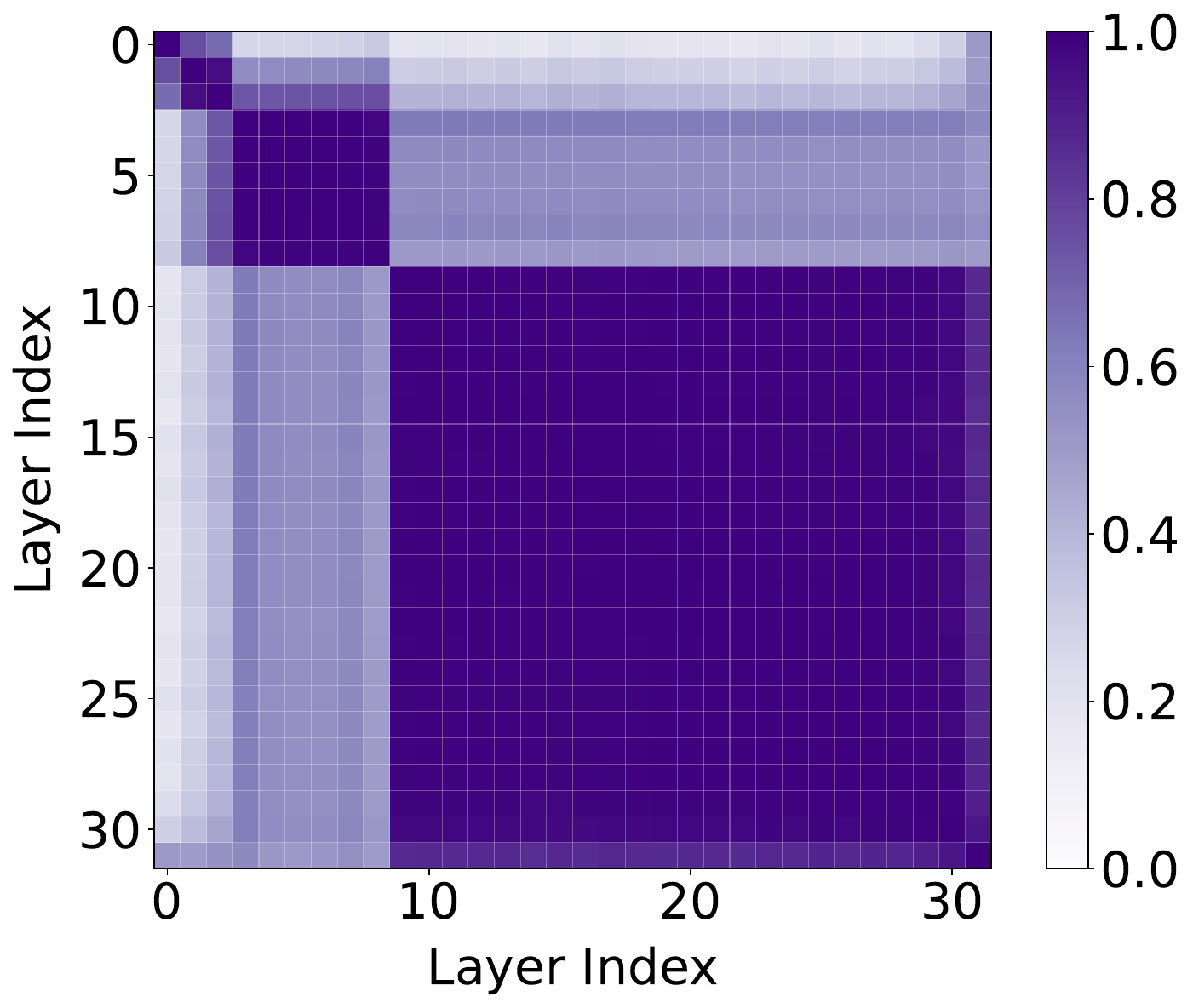}
        \caption{The attention weight cosine similarity.}
        \label{motivation:attention_similarity}
    \end{minipage}
    \hfill
    \begin{minipage}{0.73\linewidth}  
        \centering
        \begin{subfigure}{0.30\linewidth}
            \centering
            \includegraphics[width=0.9\linewidth]{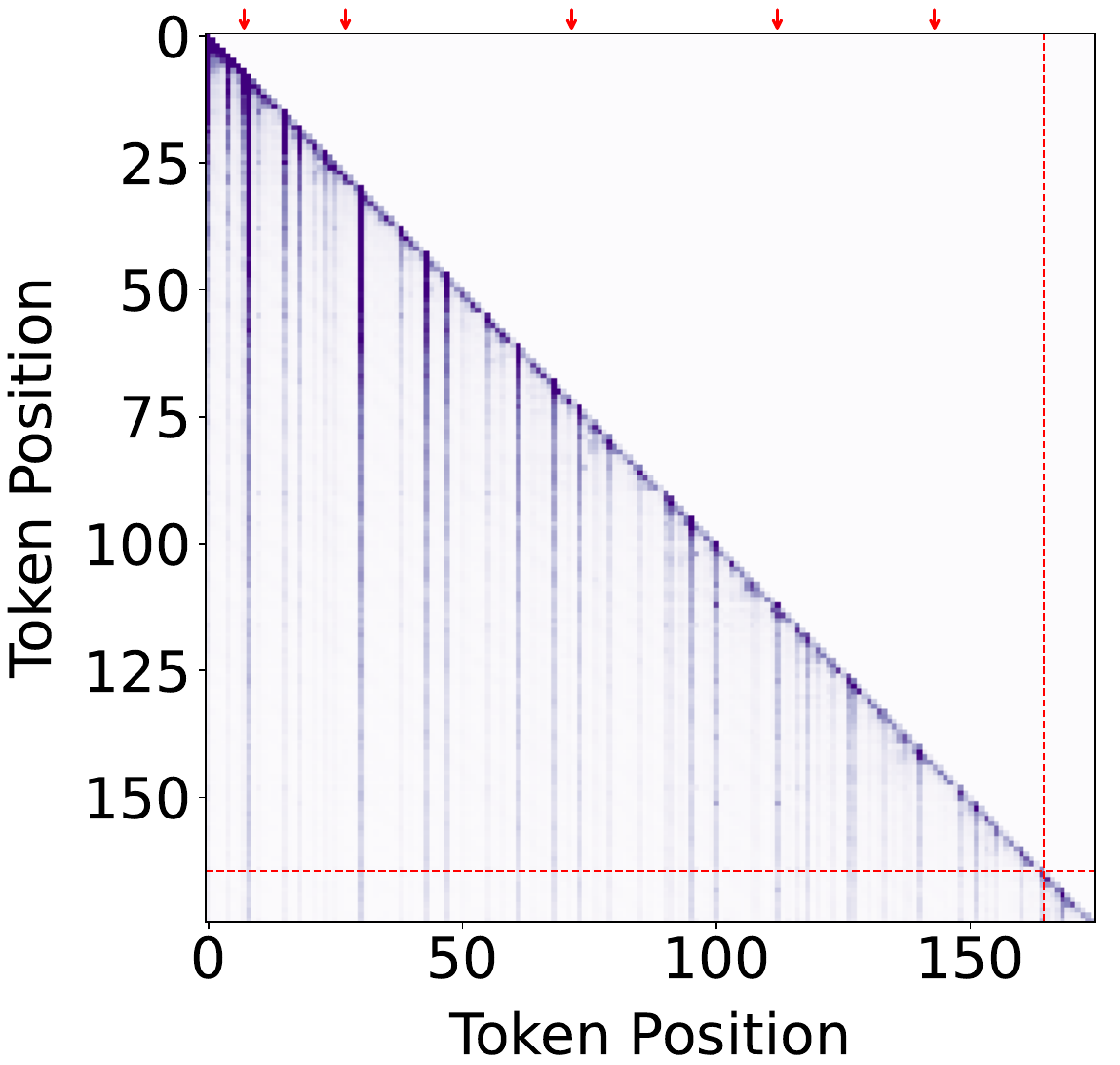}
            \caption{layer 0-2 (I-P layer)}
            \label{preexp_layer1}
        \end{subfigure}
        \hfill
        \begin{subfigure}{0.30\linewidth}
            \centering
            \includegraphics[width=0.9\linewidth]{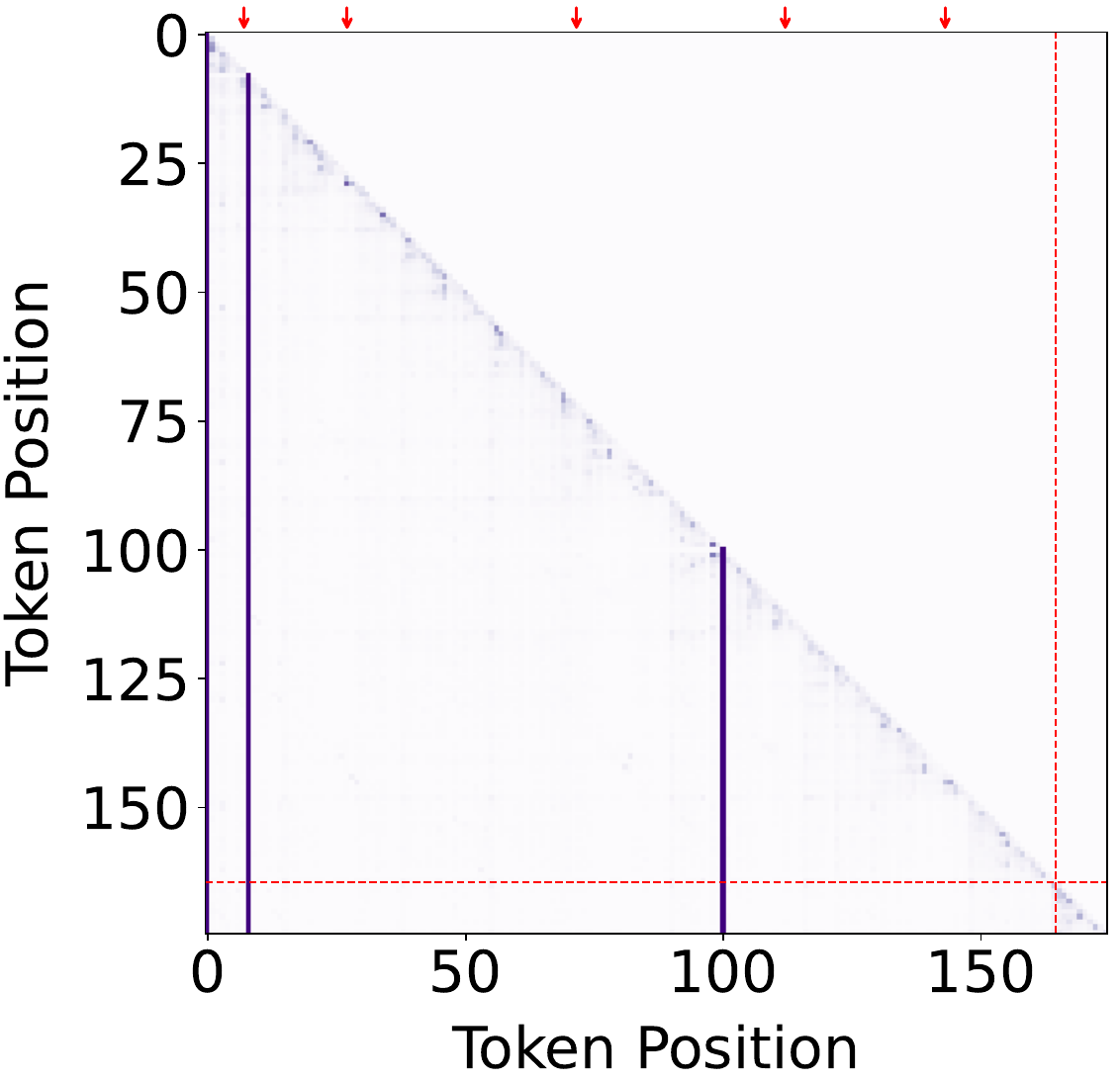}
            \caption{layer 3-8 (I-E layer)}
            \label{preexp_layer7}
        \end{subfigure}
        \hfill
        \begin{subfigure}{0.36\linewidth}
            \centering
            \includegraphics[width=0.9\linewidth]{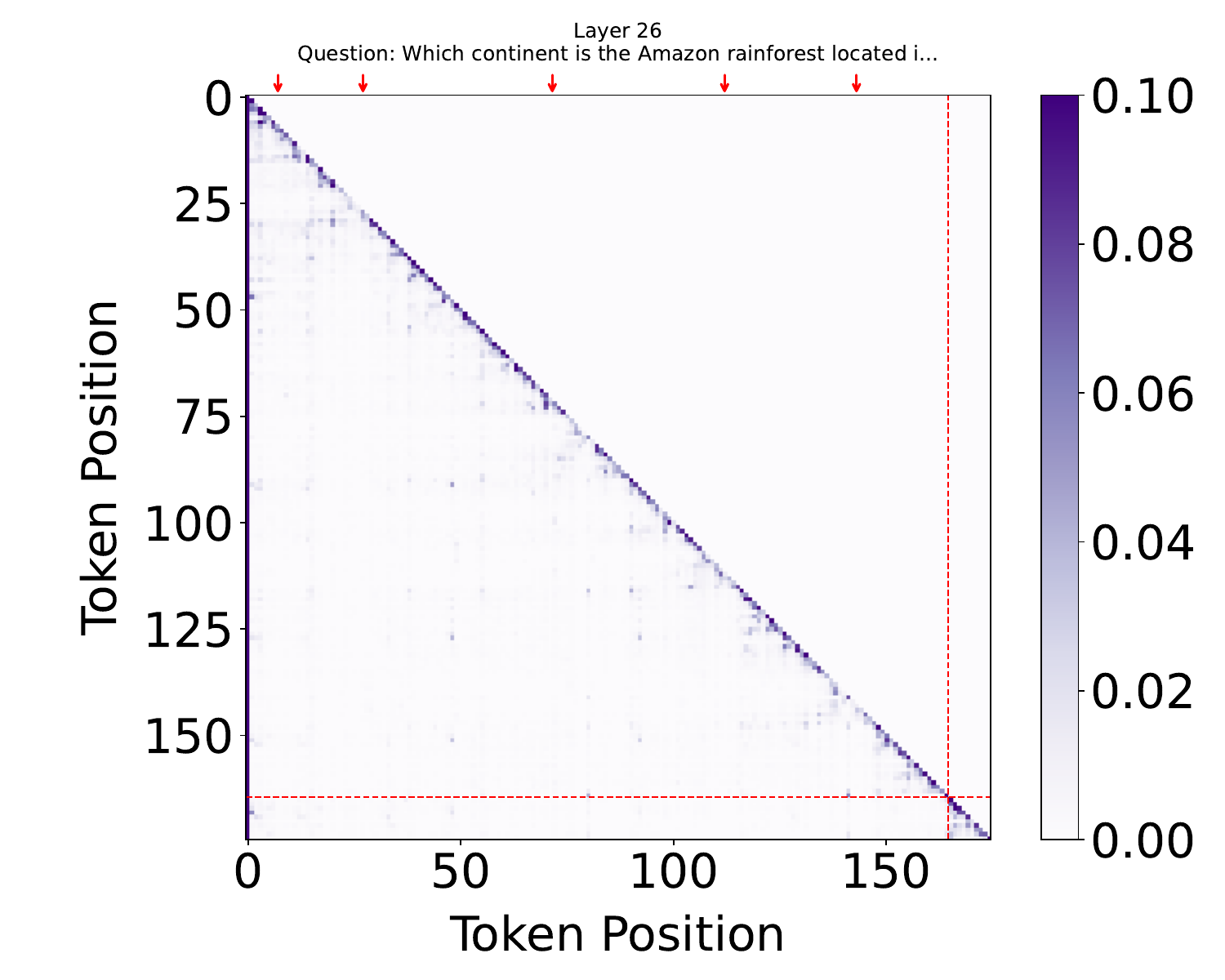}
            \caption{layer 9-31 (I-R layer)}
            \label{preexp_layer12}
        \end{subfigure}
        \caption{Three attention weight patterns. The red line represents where the question starts. Red arrows mark the position of the 5 key information.}
        \label{attention_weight_map}
    \end{minipage}
    \vspace{-8pt}
\end{figure*}

\textbf{The prefilling phase.} When a user inputs a request prompt, the model computes a series of KVs for the prompt token list $X=[x_1, x_2, \cdots, x_s]$ and generates the next token $x_{s+1}$. The KVs are stored as KV caches for use in the decoding phase.

\textbf{The decoding phase.} The decoding phase iteratively generates tokens. The decoding phase takes the hidden state of the new token $x_{s+1}$ and the KV caches of $x_{1-s}$ as input to compute the KV caches of $x_{s+1}$ and the next new token $x_{s+2}$. The decoding phase iteratively continues until the generated new token is <\textit{EOS}> or the number of generated new tokens reaches the maximum allowed token generation number. 

Modern LLM engages users in multi-turn conversations. A multi-turn conversation consists of a series of continuous conversations. In each conversation turn, a user inputs a new request or command $r_i$ and waits for the answer response $a_i$ from the model. The model generates $a_{n+1}$ by incorporating the historical tokens from all prior conversation turns $[r_1,a_1,r_2,a_2,\cdots,r_{n},a_{n},r_{n+1}]$ to sustain a cohesive and intelligible conversation flow.

To have a comprehensive understanding of the multi-turn conversations in daily life, we take an analysis of ShareGPT~\cite{chen2024sharegpt4v}, a real dataset collected from ChatGPT that includes more than 90K conversations. We calculate that the average length of each conversation is 630 tokens, as shown in Figure \ref{token_turn}, which is a relatively high increasing speed. Originally, the model evicts the KV caches of historical conversations at the end of a conversation turn. As a result,  with the increase of conversation turns, the prefilling cost becomes the main cost of the whole model inference process, as shown in Figure \ref{prefill_pingjing}.

Since the KV caches loading cost from CPU to GPU is less than the recomputation cost of the whole historical conversation as shown in Figure \ref{prefill_pingjing}, existing works offloads the whole KV caches to CPU when the conversation is inactive and fetch the target KV caches when the conversation receives new user inputs to jump over the prefilling phase of historical conversations. However, this approach still has two identical overheads:

\begin{figure}[t]
	\setlength\abovecaptionskip{0.2\baselineskip}
	\setlength\belowcaptionskip{-0.5\baselineskip}
	\includegraphics[width=\linewidth]{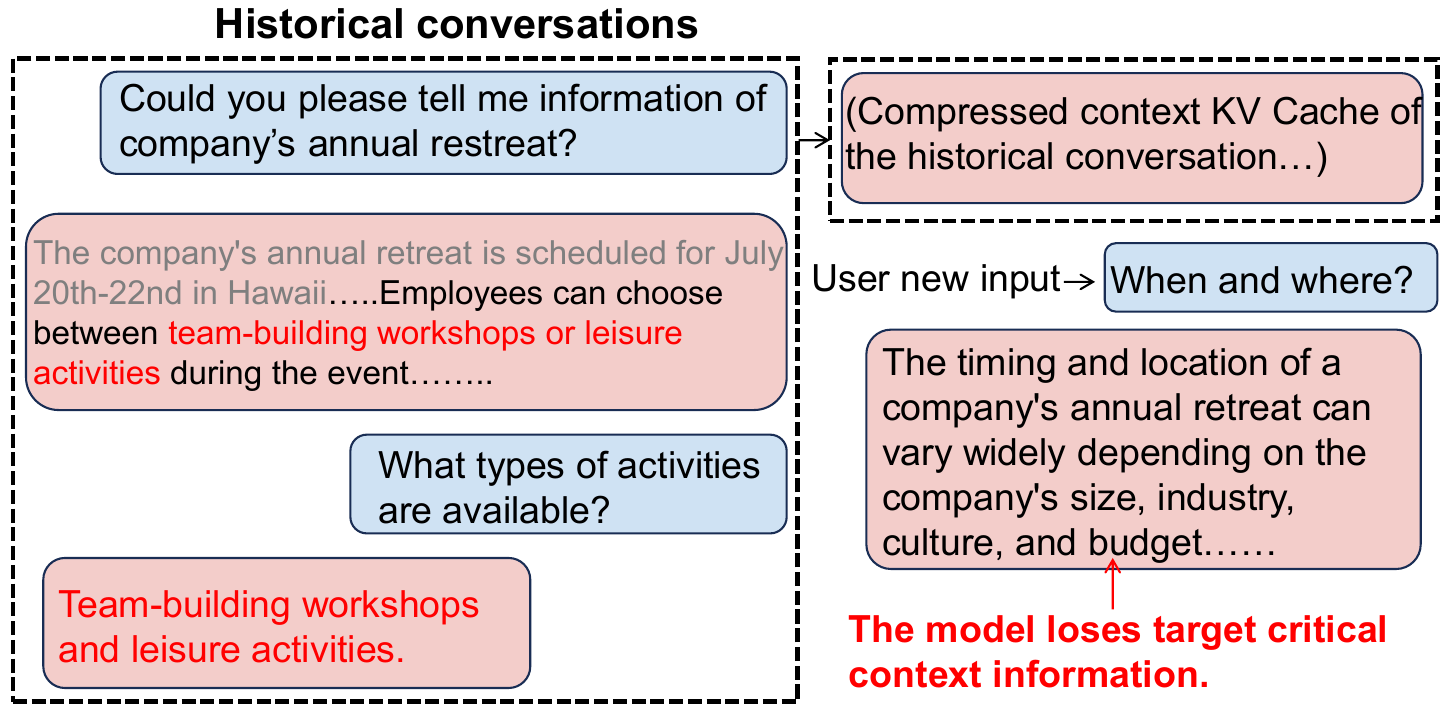}
	\caption{An example of information damage due to KV caches compression of multi-turn conversation. }
	\label{Motivation:compress}
\end{figure}

\begin{itemize}
    \item \textbf{Storage overhead.} The KV caches of inactive conversations take a large amount of memory space, as shown in Figure \ref{kv_storage}.  For example, Qwen1.5-7B takes an average of 1GB of KV cache space for each conversation turn with 2000 tokens. As a result, around 50 multi-turn conversations will soon take up the memory space of the CPU.  
    \item \textbf{Loading overhead.} If the CPU space is full, the model offloads the inactive KV caches from DRAM to SSD to release the memory of the CPU. The loading cost from SSD is further larger than from DRAM, as shown in Figure \ref{KV_load}.
\end{itemize}

\subsection{Motivation and Challenges}

\label{strawman}

To reduce the storage and loading overhead associated with state restoration, prior works~\cite{liu2024minicache, yang2024kvsharer, mu2024cross} have explored KV cache compression for inactive conversations. They compress KV caches across adjacent layers with highly similar attention
patterns. However, these methods typically employ a fixed compression strategy that most commonly compresses KV caches from every two adjacent deep layers, without considering the specific conversation context, which often degrades generation quality. To investigate this limitation, we conduct an attention weight similarity analysis, as illustrated in Figure~\ref{motivation:attention_similarity}. Our findings reveal that high attention similarity is not limited to adjacent deep layers but is also present across distant and even shallow layers. This insight suggests a broader, more flexible compression strategy selection space. To explore whether other compression strategies can still maintain the generation quality, we generate three randomized compression strategies where compression is applied across non-adjacent and shallow layers. We evaluate their impact on generation accuracy using the LongBench benchmark~\cite{bai2024longbench2}, with results presented in Table~\ref{fix_accuracy}. The results show that no single strategy consistently outperforms others across all conversations—indicating that the optimal compression strategy is conversation-dependent. 

Motivated by this, we consider a compression approach that applies conversation-adaptive KV cache compression based on attention weight similarity. However, this approach introduces three significant challenges.


\begin{table}[t]
\caption{The accuracy of three compression strategies.}
\label{fix_accuracy}
\begin{tabular}{@{}ccccc@{}}
\toprule
             & gov\_report    & vcsum          & lsht          & passretriev\_zh \\ \midrule
baseline & 25.3           & 16.44          & 19.5          & 10              \\
strategy1    & 20.06          & 15.96          & \textbf{20.5} & 6               \\
strategy2    & \textbf{23.24} & 16.02          & 18.3          & \textbf{9.5}    \\
strategy3    & 19.22          & \textbf{16.21} & 19.2          & 8               \\ \bottomrule
\end{tabular}
\end{table}

\begin{figure}[t]
    \setlength\belowcaptionskip{-0.5\baselineskip}
    \centering
    \begin{minipage}{0.47\linewidth}
        \includegraphics[width=\linewidth]{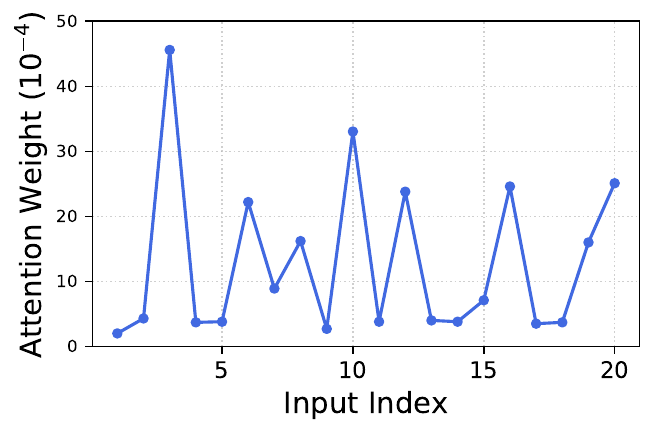}
        \caption{Average attention weight of a single token varying different user inputs.}
        \label{attn_vary}
    \end{minipage}
    \hfill
    \begin{minipage}{0.49\linewidth}
    \includegraphics[width=\linewidth]{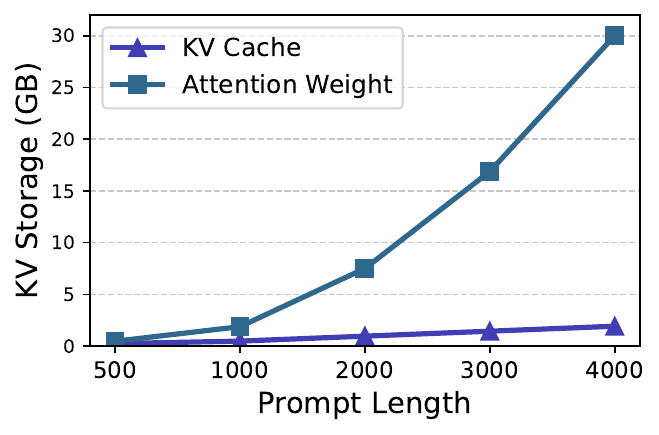}
        \caption{A comparison of KV caches and attention weights storage space varying different prompt lengths.}
        \label{KV-attn}
    \end{minipage}
\end{figure}

\textbf{Challenge 1: Compression without Knowledge of Future User Input.} 
Our compression approach leverages attention weights to quantify token-level importance, enabling selective compression of KV caches corresponding to less significant tokens. However, in multi-turn conversations, the relative importance of individual tokens dynamically shifts across different user inputs. As illustrated in Figure~\ref{Motivation:compress}, tokens that appear unimportant in the current conversation (i.e., those with low attention weights) may later become crucial in responding to a future user query. We generate a context and 20 different inputs to test the attention weight variance as shown in Figure \ref{attn_vary}. We observe that the attention weights assigned to the same token can \textbf{fluctuate significantly across different user inputs}. Over-aggressive compression risks permanently discarding context that could be semantically or logically relevant in later turns. The central challenge is, therefore, how to compress historical KV caches in a way that minimizes memory and loading overhead while preserving the model’s ability to respond accurately and coherently to all possible future user inputs.


\textbf{Challenge 2: Overhead Introduced by the Compression Process.}
To enable conversation-adaptive compression, we compute similarity between attention weight patterns across model layers, identifying which layers can be jointly compressed. However, this process introduces considerable overhead. Specifically, attention weights have a shape of \texttt{[batch\_size, num\_heads, seq\_len, seq\_len]}, which scales quadratically with sequence length. In contrast, KV caches have a shape of \texttt{[batch\_size, num\_heads, seq\_len, head\_dim]}, where \texttt{head\_dim} is typically much smaller (e.g., 128). We test the storage space of KV caches and attention weights using Qwen1.5-7B as shown in Figure \ref{KV-attn}. We observe that the storage space of attention weight increases sharply with the increase of prompt length compared with the KV caches. This discrepancy becomes especially problematic as the number of tokens grows, making it infeasible to store the full attention weight for long conversations without significantly increasing memory consumption. Moreover, computing attention similarity across all layers adds further latency and computation overhead. Thus,  we must balance compression effectiveness with the real-time constraints of inference workloads to make the system practical.

\begin{figure}
\centering
    \begin{subfigure}{1.0\linewidth}
        \centering
        \includegraphics[width=1.0\linewidth]{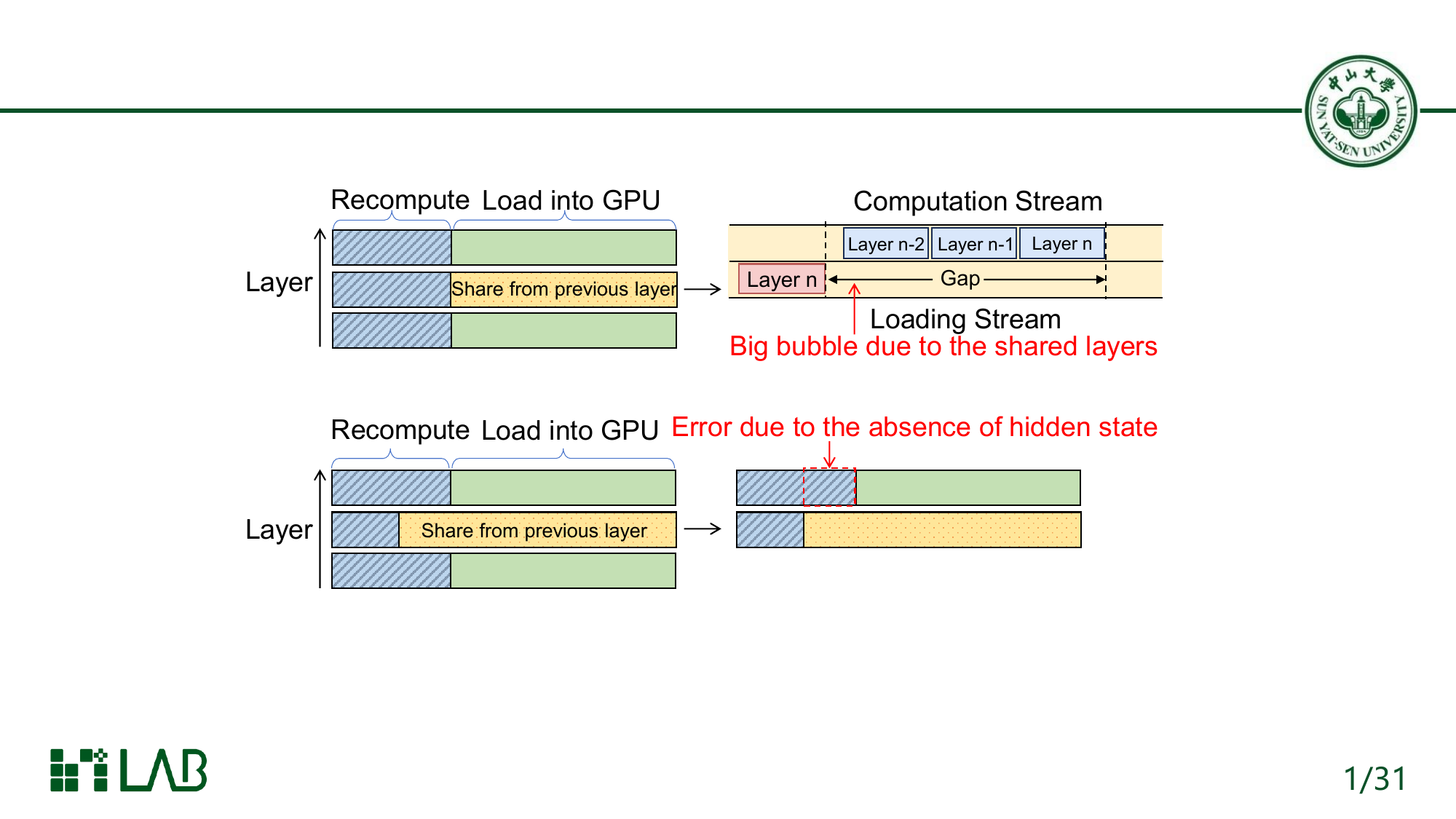}
        \caption{Big bubble due to the shared layers need not load KV caches.}
        \label{pipeline_gap}
    \end{subfigure}
    \hfill
    \begin{subfigure}{1.0\linewidth}
        \centering
    \includegraphics[width=1.0\linewidth]{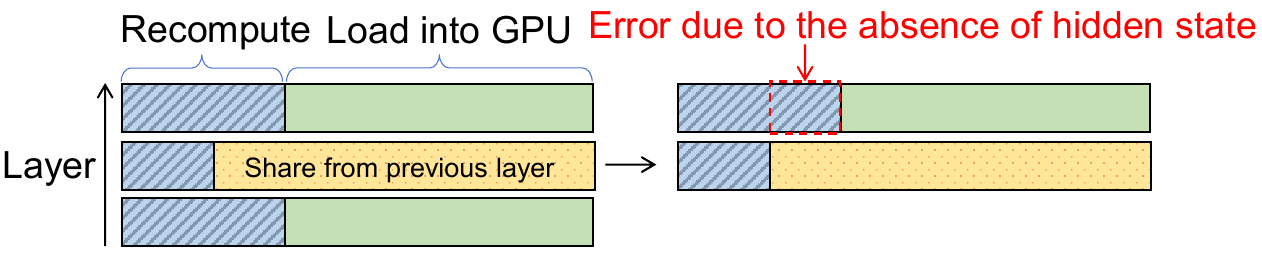}
        \caption{Prefiling error due to the absence of the related hidden state.}
        \label{pipeline_error}
    \end{subfigure}
    \caption{Two possible performance degradation problems in the prefilling computation-loading pipeline.}
    \label{prefilling_problem}
    \vspace{-8pt}
\end{figure}

\textbf{Challenge 3: Efficient Restoration Using Compressed KV Caches.}
State restoration in modern inference systems often employs a recomputation-loading pipeline~\cite{yu2025stateful, jin2024compute, yin2024llm}, which overlaps computation and I/O to minimize latency. By simultaneously recomputing certain tokens while loading others, this hybrid pipeline achieves better throughput than pure recomputation or loading alone. However, this efficiency relies on a predictable cache layout. As shown in Figure \ref{pipeline_gap}, when KV caches are compressed, the original balance between recomputation and loading is disrupted. Certain layers or tokens may require more recomputation, while others depend heavily on loading, leading to idle periods or "bubbles" in one of the streams. In addition, as shown in Figure \ref{pipeline_error}, how to divide the recomputation, the loading, and the sharing part needs to be defined properly. Since loading or sharing will discard the corresponding hidden states, improper division may lead to a prefilling error due to the absence of hidden states. Therefore, a new pipeline scheduling mechanism is required to account for the altered cache distribution and reestablish efficient overlap between loading and computation.

\section{Overview of Krul}

\begin{figure*}[t]
	\setlength\belowcaptionskip{-0.5\baselineskip}
	\includegraphics[width=0.9\linewidth]{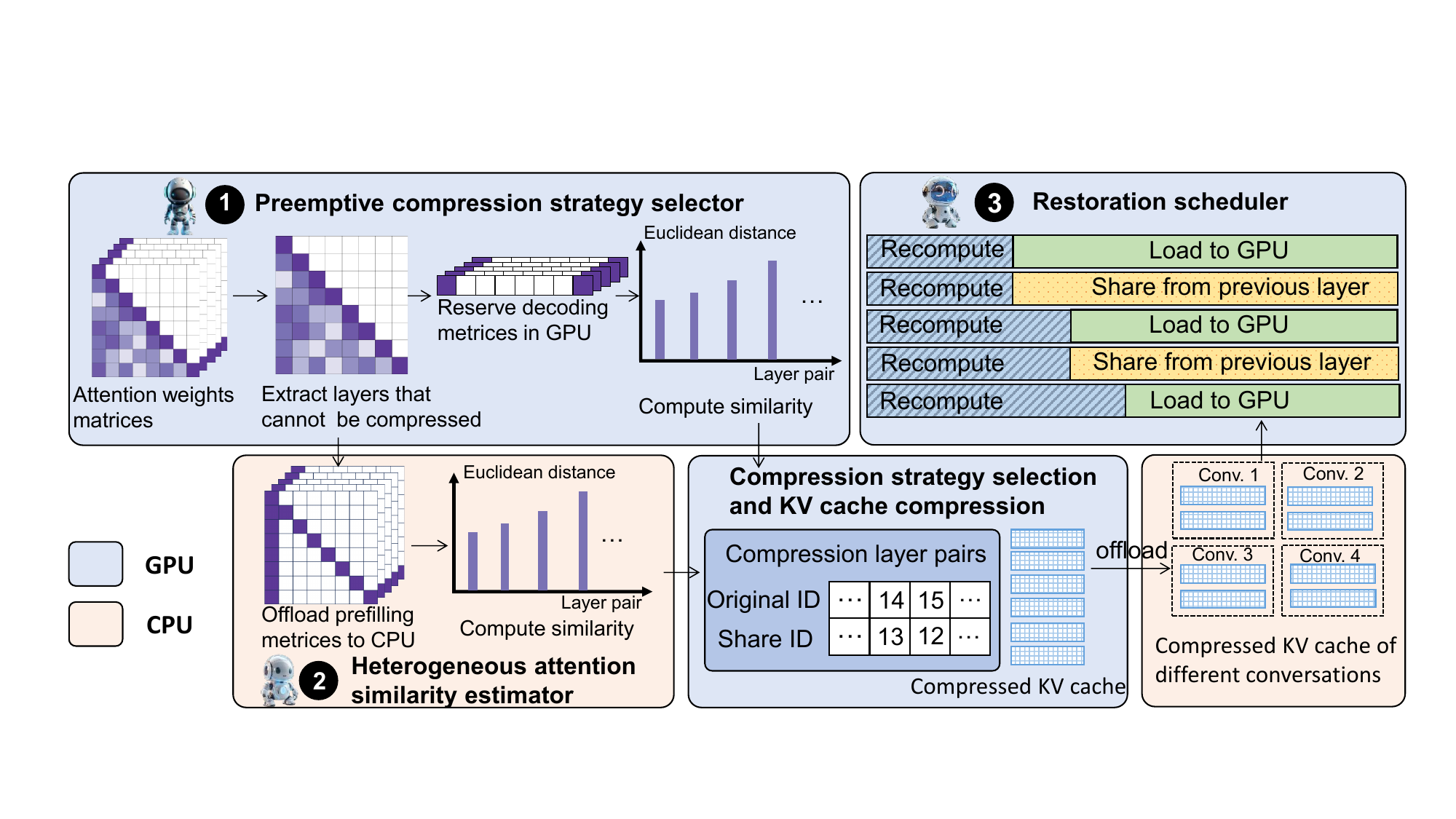}
	\caption{The system architecture of Krul.}
	\label{system-overview}
\end{figure*}

In this paper, we present Krul, an efficient inference system for LLMs that employs conversation-adaptive KV cache compression to enable fast and memory-efficient state restoration without compromising generation quality. An overview of Krul's architecture is shown in Figure~\ref{system-overview}. When a conversation ends and becomes inactive, Krul compresses the associated KV caches and offloads them to CPU memory. Upon reactivation of the conversation, Krul restores the model state by parallelizing recomputation and loading, significantly reducing both computational overhead and memory usage. Compared to the state-of-the-art restoration inference systems, Krul achieves substantial improvements in both performance and resource efficiency.

For Challenge 1, to prevent degradation in generation quality, we design the preemptive compression strategy selector. This module analyzes the distribution of attention weights and identifies model layers input-sensitive model layers, and thus unsuitable for compression. After extracting those layers, it generates conversation-adaptive compression strategies optimized to minimize accuracy loss across diverse conversations.

For Challenge 2, to mitigate the additional storage and computation costs introduced by adaptive compression, we propose the token-wise heterogeneous attention similarity estimator. We analyze the feature of conversations and find a similarity estimation memory-time trade-off of the prefilling phase and the decoding phase. Based on this observation, we apply a token-wise partition to the attention weights and use CPU and GPU to heterogeneously compute the similarity of attention weights.  At the end of each conversation, we aggregate the similarity results to guide KV cache compression and offload to CPU memory.

For Challenge 3, to overcome imbalances between recomputation and loading caused by KV caches compression, we introduce the bubble-free restoration scheduler. This module includes a hardware-aware optimization algorithm that automatically tunes compression and scheduling parameters to suit different CPU-GPU performance configurations. Moreover, this module dynamically orchestrates the recomputation and loading ratio across layers to ensure full pipeline overlap, thereby minimizing the latency. It also guarantees correctness by detecting and resolving any recomputation gaps that may arise due to missing hidden states from compression. 

\section{Preemptive Compression Strategy selector}

\begin{table*}[t]
\caption{The attention weight of five questions to five keys~($10^{-4}$)}
\label{motivation:attention_weight_table}
\centering
\setlength{\tabcolsep}{7pt} 
\renewcommand{\arraystretch}{0.8} 
\begin{tabular}{@{}c *{15}{c}@{}}
\toprule
question\_id & \multicolumn{3}{c}{question 1} & \multicolumn{3}{c}{question 2} & \multicolumn{3}{c}{question 3} & \multicolumn{3}{c}{question 4} & \multicolumn{3}{c}{question 5} \\ 
\cmidrule(lr){2-4} \cmidrule(lr){5-7} \cmidrule(lr){8-10} \cmidrule(lr){11-13} \cmidrule(lr){14-16}
layer\_id & 2 & 4 & 26 & 2 & 4 & 26 & 2 & 4 & 26 & 2 & 4 & 26 & 2 & 4 & 26 \\ 
\midrule
key1 & {\color{red} 16.6} & {\color{red} 5.0} & {\color{red} 45.6} & 15.3 & 4.3 & {\color{red} 44.5} & 14.8 & 4.2 & {\color{red}14.0} & 15.2 & 4 &   {\color{red}23.8} & 16 & 4.2 & {\color{red} 33.05} \\ 
key2 & 21.9 & 3.5 & 9.5 & {\color{red} 24.6} & {\color{red} 8.9} & 24.1 & 22.3 & 3.4 & 4.4 & 21.2 & 3.1 & 5.4 & 21.5 & 3.3 & 13.8 \\ 
key3 & 17.1 & 4.0 & 5.4 & 16.2 & 4.0 & 20.8 & {\color{red} 23.4} & {\color{red} 4.5} & 12.4 & 16.4 & 3.7 & 22.2 & 16.4 & 4.1 & 14.5 \\ 
key4 & 61.9 & 3.7 & 1.8 & 56.9 & 3.6 & 10.3 & 59 & 3.8 & 3.4 & {\color{red} 62.4} & {\color{red} 4.4} & 7.2 & 52.7 & 2.4 & 7.1 \\ 
key5 & 26.7 & 2.8 & 6.2 & 26.7 & 5.1 & 12.7 & 25.1 & 2.7 & 5.2 & 25 & 2.6 & 3.7 & {\color{red} 26.8} & {\color{red} 4.6} & 14.4 \\ 
\bottomrule
\end{tabular}
\end{table*}

\subsection{Attention Pattern Analysis}

To investigate how attention patterns vary across different user prompts for the same conversation history, we design a controlled experiment to simulate multi-turn conversation scenarios. Specifically, we construct a context sequence containing five pieces of key information, evenly distributed throughout the sequence, representing a historical conversation. We then formulate five corresponding user queries, each targeting one specific piece of information, to simulate different user inputs. Each query is concatenated with the full context and passed independently into the \texttt{Qwen1.5-7B} model, resulting in five distinct input sequences.

For each input, we extract and visualize the attention weights from selected layers, as illustrated in Figure~\ref{attention_weight_map}. These visualizations reveal clear differences in how various layers attend to the context based on the user query. To further quantify these observations, we analyze attention weights in detail across layers 2, 4, and 26—chosen to represent distinct attention behaviors—and summarize the results in Table~\ref{motivation:attention_weight_table}. Our analysis reveals three distinct attention patterns:

\begin{table}[t]
\caption{The I-R layer analysis}
\label{motivation:I-R-layer}
\begin{tabular}{@{}c@{\hspace{0.5em}}c@{\hspace{0.5em}}c@{\hspace{0.5em}}c@{\hspace{0.5em}}c@{}}
\toprule
              & Deepseek-6.7B & LLaMA-13B & Qwen2-72B \\ \midrule
layer num     & 32       & 40             & 80       \\
I-R layer num & 24       & 32             & 70       \\
ratio         & 75\%     & 80\%        & 87.5\%   \\ \bottomrule
\end{tabular}
\end{table}

\begin{itemize}
\item \textbf{Information-Extraction (I-E) Layers:} In certain mid-level layers (e.g., layer 4), we observe that each question strongly attends to the specific context token containing the relevant information. This behavior suggests that the model is selectively extracting task-relevant information based on the user input, indicative of a targeted extraction mechanism.
\item \textbf{Information-Processing (I-P) Layers:} In shallow layers (e.g., layer 2), although attention is more distributed, we find that the same key token often receives the highest attention from its corresponding question across different inputs. These layers appear to perform holistic sentence-level processing influenced by user prompts, adapting the attention structure according to the input.
\item \textbf{Initial-Recent (I-R) Layers:} In deeper layers (e.g., layer 26), attention weights are largely invariant across different user inputs. The model consistently assigns the highest weights to the first token and shows elevated attention toward tokens near the end of the sequence, regardless of query content. This pattern indicates a fixed bias toward the initial and most recent context, with little sensitivity to input prompt variation.
\end{itemize}

We extend this analysis to additional models—\texttt{Deepseek-6.7B}, \texttt{LLaMA-13B}, and \texttt{Qwen2-72B}—to assess the generality of these patterns. As shown in Table~\ref{motivation:I-R-layer}, we observe that I-R layers constitute a substantial proportion of total layers across all models studied. This prevalence suggests that a large portion of layers in modern LLMs exhibit input-insensitive attention behaviors, which has direct implications for designing more effective and safe compression strategies.

\subsection{Compression Strategy Selection}

\textbf{Non-I-R Layer Extraction}. Given that a substantial portion of model layers exhibit the I-R attention pattern—where attention distributions remain largely invariant to user input—these layers present a promising opportunity for safe and aggressive KV cache compression. Since I-R layers primarily focus on the beginning and end of the conversation, regardless of the user input, compressing their KV caches is unlikely to degrade generation quality. However, to preserve semantic fidelity, it is critical to first identify and exclude layers that do not follow the I-R pattern.   

We introduce an efficient, lightweight online method to detect non-I-R layers based on attention weight distribution, as outlined in Algorithm~\ref{algorithm:optimizer_step} lines 1$\sim$8. For a given sequence of length $s$, we define the initial token subset as $T_{initial}=T[:10\%\times s]$ and the recent token subset as $T_{recent}=T[90\%\times s:]$. For each layer $l_i$, we compute the average attention weight directed toward these two regions across all heads and positions.  If the combined average attention to $T_{initial}\cup T_{recent}$ does not exceed a predefined threshold $\gamma$, we classify the layer as non-I-R and exclude it from compression to avoid compromising the model's responsiveness to user-specific inputs. This identification process ensures that compression is applied selectively and safely, preserving the model's ability to generate accurate and context-aware responses for multi-turn conversations.

\begin{algorithm}[t]
\DontPrintSemicolon
\caption{Preemptive Compression Strategy Selection}
\label{algorithm:optimizer_step}
\KwIn{Model $\mathcal{F}$, prompt $\mathcal{C}$, attention weight threshold $\gamma$, model layers $\mathcal{L}$, compression ratio $r_l$}
\KwOut{KV compression layer pairs $\mathcal{M}$}
Initialize attention weights $\mathcal{A}$\;
$\mathcal{A}=\mathcal{F}.forward(\mathcal{C}).attention\_outputs$\;

\For{each layer $i$}{
        \textbf{EXTRACT} $T_{initial}=T[:10\%\times seq\_length]$\;
        $T_{recent}=T[90\%\times seq\_length:]$\;
        $avg\_weight\_sum=\frac{\sum(\sum A_i(T_{initial})+\sum A_{i}(T_{recent}))}{seq\_length}$\;
        \If {$avg\_weight\_sum < \gamma$}{
            $\mathcal{L}_N=\mathcal{L}_N\cup i$\;
        }
}
$\mathcal{L}_I=\mathcal{L}- \mathcal{L}_{N}$\;
Initialize attention distance $\mathcal{D}=\varnothing$, $N=layer\_num$\;
\For{ every two weights $A_i, A_j(i>j) \in \mathcal{A}$ \textbf{and} $i, j \in \mathcal{L}_I$ }{
    $D_{i,j}=Euclidean\_Distance(A_i,A_j)$ \;
    $\mathcal{D}=\mathcal{D} \cup D_{i,j}$\;
}
Initialize shared layers $\mathcal{S}=\varnothing$, $\mathcal{M}=\varnothing$\;
\For{ each $D_{i,j}$ in $\mathcal{D}$}{
\If {$i, j \notin \mathcal{S}$}{
    $\mathcal{S}=\mathcal{S}\cup\{i,j\}$\;
    $\mathcal{M}=\mathcal{M}\cup\{(i,j)\}$\;
    }
        \Else {
            \textbf{continue}\;
    }
    \If {$|\mathcal{S}| \geqslant N \times r_l$}{
    \textbf{break}\;
    }
}
\end{algorithm}

\textbf{Compression Strategy Selection for I-R Layers.}
Once non-I-R layers have been excluded from compression to preserve generation quality, we focus on compressing the KV caches in the remaining I-R layers of inactive conversations. To control the compression granularity, we define a tunable hyperparameter: the compression ratio $r_l$, which specifies the proportion of I-R layers to be compressed.

To guide the compression process, we leverage the attention weights generated during the model's forward pass. These attention weights capture how information is distributed across the sequence and layers, enabling us to identify candidate layers and token spans where KV cache sharing will minimally impact model performance. Using these attention weights, we perform a constrained search over the compression strategy space to find a configuration that satisfies the given ratio $r_l$, while minimizing expected accuracy loss. The complete strategy search and compression workflow is formalized in Algorithm~\ref{algorithm:optimizer_step}. We decompose this algorithm step-by-step and provide a detailed explanation of its operational logic and design rationale.

Firstly, we need to estimate the attention similarities across I-R layers, as outlined in Algorithm~\ref{algorithm:optimizer_step} lines 10$\sim$13. Firstly, we initialize the attention distance $\mathcal{D}$. Then we flatten the attention weight of each layer and attention head into a one-dimensional vector. Next, we calculate the Euclidean distance between the attention weights of any two layers in $\mathcal{L_I}$ and calculate the average distance to obtain $\mathcal{D}$. Then we sort $\mathcal{D}$ in ascending order as a smaller Euclidean distance indicates higher similarity. Consequently, similar layer pairs are prioritized. We then set two variables, KV compression layer pairs $\mathcal{M}$ and shared layers $\mathcal{S}$,  to record the candidate KV caches compression strategy and layers that have been shared.

Next, we search the compression strategy based on the similarity, as outlined in Algorithm~\ref{algorithm:optimizer_step} lines 14$\sim$22. Based on the values in $\mathcal{D}$, we first sequentially check whether a pair of layers $(i,j) \in \mathcal{S}$. If a layer $i$ is shared from previous layers or shared to another layer, layer $i$ needs to be jumped to avoid significant accuracy degradation. If not, we select the pair of layers $(i,j)$ to add to $\mathcal{M}$ for compression and record $(i,j)$ in $\mathcal{S}$ to disable layer $i$ and $j$ from compression. When the number of shared layers reached the layer number compression ratio $r_l$, we break the searching loop and output the compression strategy $\mathcal{M}$.

\section{Token-wise
Heterogeneous Attention Similarity Estimator.}

\label{Heterogeneous}

Since our compression algorithm is based on attention weights, it brings two additional overheads: (1) attention weight storage overhead. Because the core of our compression strategy is grounded in the similarity between attention weights, we must retain these weights during the generation process. This requirement adds a non-trivial memory burden, particularly in multi-turn conversations where attention histories can grow substantially; (2) similarity computation overhead. To perform compression, our algorithm computes the Euclidean distance between attention weights—typically at the end of the conversation. While this step is essential for accurate similarity assessment, it introduces additional computational load during or after the generation phase, potentially impacting latency and throughput. Mitigating these overheads is crucial to ensure that our compression approach remains both scalable and practical in real-world deployment scenarios. 

\begin{figure}[t]
\centering
    \begin{subfigure}{0.49\linewidth}
        \centering
        \includegraphics[width=1.0\linewidth]{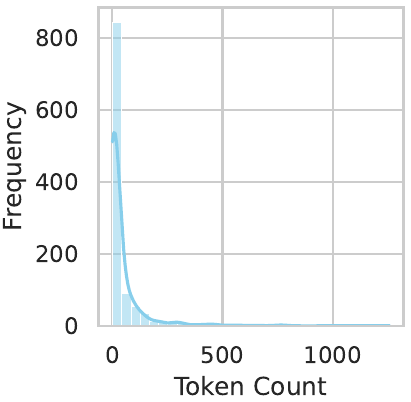}
        \caption{User input distribution.}
        \label{userinput}
    \end{subfigure}
    \hfill
    \begin{subfigure}{0.49\linewidth}
        \centering
        \includegraphics[width=1.0\linewidth]{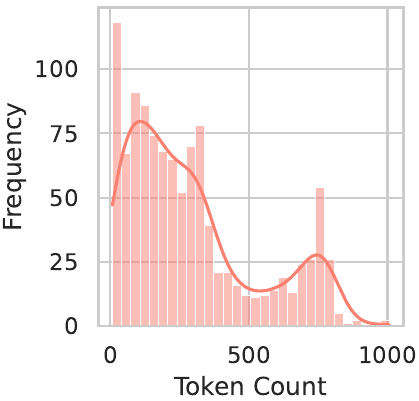}
        \caption{Model output distribution.}
        \label{model_output}
    \end{subfigure}
    \caption{The comparison of user input and model output length distribution in ShareGPT.}
    \label{comparison_user_output}
    \vspace{-8pt}
\end{figure}

We begin by analyzing the prompt lengths of user inputs and model outputs using the ShareGPT dataset, as illustrated in Figure \ref{comparison_user_output}. A clear asymmetry emerges: model outputs are significantly longer than user prompts. On average, the output sequence is approximately 6.56$\times$ longer than the corresponding user input. Let us combine this asymmetry with attention similarity computation. Specifically, we adopt Euclidean distance to quantify the similarity between attention weights. This metric choice allows us to avoid storing the entire attention history and performing post hoc computation after generation. Instead, we incrementally compute attention similarity during the forward pass of the model. 

The forward pass can be logically divided into two distinct phases: the prefilling phase and the decoding phase.  During the prefilling phase, the model attends to all preceding tokens in the input sequence, leading to the formation of large attention weights with the shape \texttt{[batch\_size, num\_heads, seq\_len, seq\_len]}. In contrast, during the decoding phase, attention is computed only for the newly generated token at each step. Consequently, the attention weights are much smaller—of shape \texttt{[batch\_size, num\_heads, 1, seq\_len]}.  The space of attention weights of the prefilling phase grows quadratically with prompt length, but only needs one similarity computation process. The space of attention weights of the decoding phase grows linearly with prompt length, but the computation process is repeated for each output token, resulting in a large cumulative computation cost.

This asymmetry leads to a trade-off: the prefilling attention weights are memory-intensive but fast to compute similarity, while decoding attention weights are lightweight per step but computationally repetitive across long sequences. To balance this trade-off, Krul employs a token-wise partitioning strategy combined with heterogeneous computation. The attention weights are partitioned along the token dimension into two parts based on the phase.

For the prefilling phase, Krul asynchronously offloads attention weights to the CPU during the forward pass. While the GPU continues with decoding, the CPU concurrently computes the similarity of the attention weights. This pipelined strategy ensures that data transfer and computation are effectively overlapped, mitigating both latency overhead and memory bottlenecks.

\begin{figure}[t]
    \setlength\belowcaptionskip{-0.5\baselineskip}
    \includegraphics[width=\linewidth]{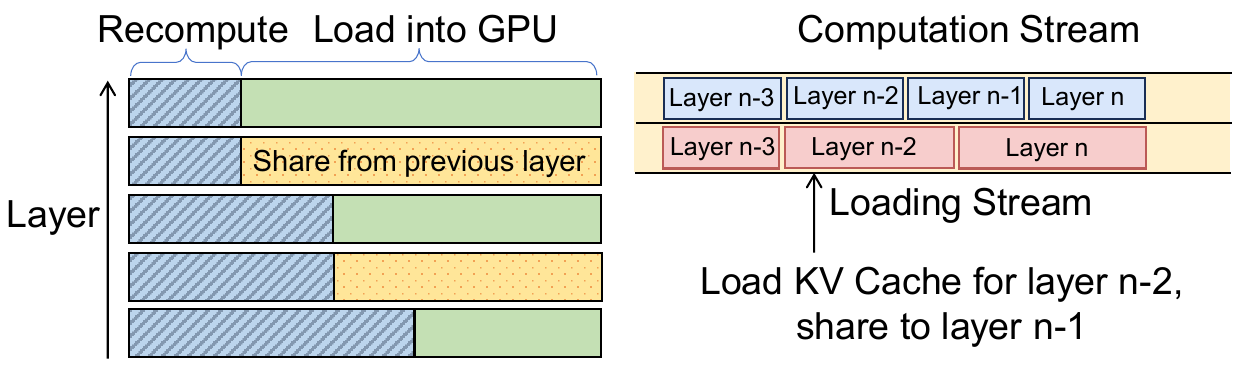}
    \caption{Example of a bubble-free restoration pipeline.}
    \label{pyramid_restore}
\end{figure}

For the decoding phase, the entire attention weight is partitioned on a per-token basis, storing only the attention weights for the most recent token. Then computes similarity with relevant layer pairs using Euclidean distance, and discards the weights immediately at the end of each decoding step. This ensures a minimal memory footprint throughout decoding. Given that attention weights are significantly smaller (approximately 256$\times$ less memory than the key-value cache, which has shape \texttt{[batch\_size, num\_heads, seq\_len, head\_dim]}, where \texttt{head\_dim} usually is 128), the cost of storing these intermediate tensors is negligible. Moreover,  by dividing similarity computation across decoding steps, Krul achieves streamed similarity estimation without materially impacting the overall generation throughput (TPOT). This fine-grained, low-latency design is key for scaling similarity-based KV cache compression mechanisms in LLMs.


\section{Bubble-free restoration scheduler}

\subsection{KV Cache Space Allocation}

To maximize the efficiency of state restoration in multi-turn conversations, Krul introduces a bubble-free restoration scheduler via a dynamic recomputation-loading pipeline designed to fully overlap computation and I/O. Central to this is determining the optimal recomputation ratio $r_c$, which balances the amount of KV cache that should be recomputed versus loaded. Since hardware capabilities vary across inference environments, we formulate this as an optimization problem:

\begin{equation}\label{eq:opt_problem}
    \begin{aligned}
        \min \quad & \lvert T_C-T_L \rvert \\
        \text{s.t.} \quad & T_C = \frac{F(N,L,d)}{F_{peak}} \\
        & T_L = \frac{G(N,L,d)}{B_{peak}} \\
    \end{aligned}
\end{equation}
 ere, $T_C$ and $T_L$ represent the recomputation and loading latency, respectively. $N$ is the number of layers, $L$ is the prompt length, $d$ is the hidden state dimension, $F(N,L, d), G(N,L, d)$ represent the total recomputation operations and loading operations workload, $F_{peak}$ and $B_{peak}$ denote the peak computation speed and memory bandwidth of the inference hardware, respectively. By minimizing the absolute difference $|T_C-T_L|$, we ensure maximum overlap between recomputation and loading, improving the overall restoration speed. Usually, the optimal value of $r_c$ can be found by preparing a calibration dataset and computing multiple model forwards with different $r_c$.

Once the optimal $r_c$ is determined, we use it-along with the compression strategy pairs $\mathcal{M}$-to allocate KV cache storage space across layers. Let $c$ be the storage space per layer for the entire conversation. We allocate total cache space as $C=\{c\times (1-r_c) \}\times N$. If a layer $l_i$ shares its KV cache with a preceding layer as defined in $\mathcal{M}$, its allocated storage $c_i$  is divided among the deep layers accordingly. We then compress the KV caches for each layer pair $(i,j)\in\mathcal{M}$, followed by offloading the compressed KV caches to CPU memory\footnote{Our compression method follows Minicache.}.

\subsection{State Restoration}

Upon reactivation of a conversation, Krul restores the model via a recomputation-loading pipeline scheduler, where computation operations recompute layer-wise KV caches while loading operations fetch KV caches from DRAM to HBM in parallel.

The key insight behind this scheduler is that the distribution of recomputation and loading volume across layers follows a pyramid-shaped pattern, as shown in Figure \ref{pyramid_restore}. Specifically, the volume of KV caches loaded from memory $S_{load}(l)$ increases linearly with layer index, while the amount that requires recomputation $S_{comp}(l)$. decreases linearly. This preplanned distribution—defined during compression—enables the executor to:
\begin{enumerate}
\item \textbf{Optimize Resource Utilization.}
By allocating a greater share of loading tasks to deeper layers—where shared KV caches are reused—Krul maximizes memory bandwidth utilization. Formally, the loading utilization per layer is:\begin{equation}
    \eta_{\text{load}}(l) = \frac{S_{\text{load}}(l)}{B_{\text{load}}}
    \end{equation}
where $B_{\text{load}}$ is the available memory bandwidth. Meanwhile, computation overhead decays quadratically:
\begin{equation}
    O_{\text{comp}}(l) = \Theta\left(S_{\text{comp}}(l) \cdot d^2\right)
    \end{equation}
This ensures that both computation and I/O streams are fully leveraged across the pipeline.
\item \textbf{Ensure Computational Correctness.} The scheduler also guarantees correctness in recomputation. Because the lengths of hidden states required at deeper layers are shorter than those computed in earlier layers, we have: \begin{equation}
    \forall l_i, \quad H_{l_{i+1}}.length \leqslant H_{l_i}.length
    \end{equation}
    Thus, for any layer \( l \), the necessary hidden states \( H_l \) are always available from the union of previously computed states: \begin{equation}
H_l.length \leqslant H_i.length(i<l)
\end{equation}

This hierarchical dependency structure ensures there are no missing hidden states during restoration, maintaining the semantic integrity of the model's internal representations.
\end{enumerate}

In summary, Krul's pipeline scheduler delivers both performance and correctness guarantees by tailoring restoration to the structural properties of compressed KV caches and adapting to the capabilities of heterogeneous hardware.

\section{Implementation}

We implement Krul on top of PyTorch and Python. Krul integrates the implementation of popular LLMs such as LLaMA and Qwen based on Transformers. Krul uses several CUDA streams to move data between GPUs and the host memory to overlap the computation and data swapping. Continuous batch is enabled through experiments.

\noindent \textbf{Request scheduling:} When a request with historical conversation arrives, Krul will first use a recomputation-loading pipeline to restore the state of the historical conversation. For recomputation,   Krul cuts the hidden state in each layer according to the KV memory layer map $C$ using the interface $\mathtt{tensor[:target\_length]}$. For loading, Krul loads the target KV caches from the CPU to the GPU. At the end of the pipeline, Krul combines the KV cache generated by recomputing and loading using $\mathtt{torch.cat()}$.

\noindent \textbf{Attention weight management:}  Krul uses CUDA streams to manage the attention weight during the model generation. While processing the model forward of the prefilling phase,  Krul uses a CUDA stream to offload the attention weight to the CPU by using $\mathtt{tensor.to("cpu")}$. Then Krul computes the similarity in the CPU and records the similarity in the $\mathtt{distance_{prefill}}$. Secondly, for the decoding phase, Krul saves the newly generated attention weight, computes the similarity, records the distance in $\mathtt{distance_{decode}}$, and deletes the attention weight using the memory interface $\mathtt{del}$ and $\mathtt{torch.cuda.empty\_cache()}$ at the end of each decoding step. Finally, Krul adds up the similarities and gets the final results.  For two attention tensors $a$ and $b$ in CPU computing, we compute the Euclidean distance using $a^2+b^2-2ab$ instead of $|a-b|^2$ to get a better performance. For GPU computing, we use $\mathtt{torch.cdist}$ to get a better performance.

\noindent \textbf{KV cache management:} Krul uses a single CUDA stream to compress the KV cache at the end of model generation and offloads the KV cache to the CPU. When an inactive conversation becomes active again, Krul uses separate I/O threads to prefetch the corresponding KV cache from the CPU to the GPU.

\section{Performance Evaluation}

We evaluate Krul to answer the following questions:

\begin{itemize}
    \item How does Krul perform in terms of TTFT and accuracy compared to state-of-the-art state restoration methods?
    \item How does Krul perform with various models and context length?
    \item How does Krul perform in KV caches storage saving?
\end{itemize}

\subsection{Experimental Setup}

 \textbf{Testbeds.} All our experiments are performed on 4 NVIDIA A100 GPUs, each with 80GB HBM. The system is equipped with 128GB DRAM and 10 TB SSDs. GPUs are connected to the host via PCIe Gen 4.

\noindent\textbf{Models.} We evaluate Krul on  LLaMA-7B, 30B, Qwen1.5-7B, and Qwen2-72B to represent the variance of model type and size. Note that we apply 
8-bit model quantization to Qwen2-72B. 

\noindent\textbf{Workloads.} We use LongBench~\cite{bai2024longbench2} to evaluate the performance and quality of model output, and we use ShareGPT~\cite{chen2024sharegpt4v} to evaluate the throughput. For accuracy evaluation,  we split the last 128 tokens of the context to simulate the user's input. Then we apply KV caches compression to the remaining context to simulate the compression of the historical conversation. 

\begin{figure*}[t]
	\setlength\belowcaptionskip{-0.5\baselineskip}
	\includegraphics[width=\linewidth]{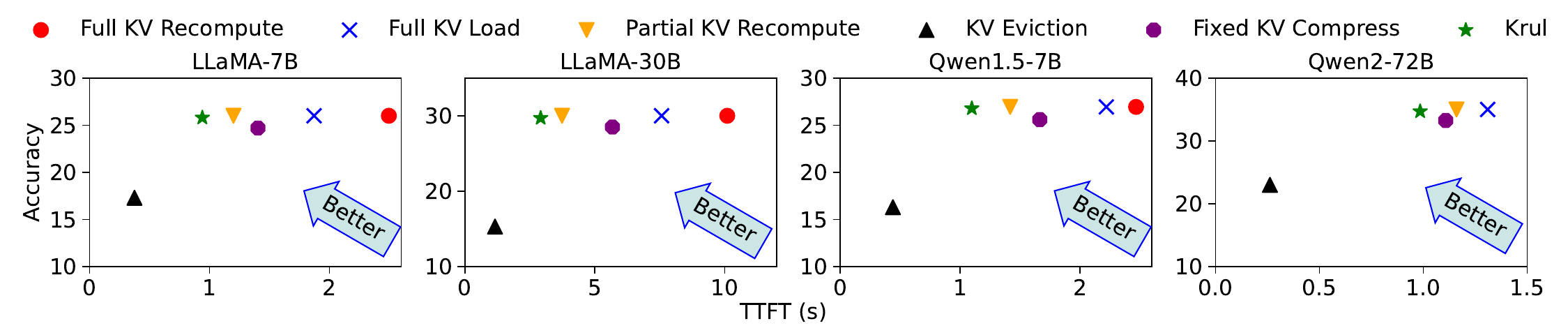}
	\caption{The overall performance comparison of Krul and other SOTA works.}
	\label{overallperformance}
\end{figure*}

\begin{figure}[t]
    \setlength\belowcaptionskip{-0.5\baselineskip}
    \centering
    \begin{minipage}{0.48\linewidth}
        \includegraphics[width=\linewidth]{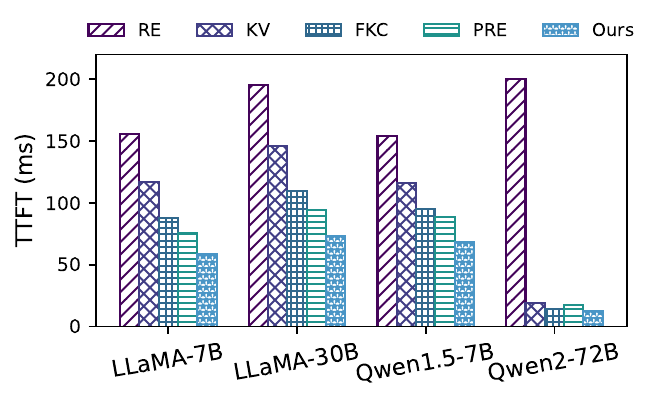}
        \caption{Time to first token.}
        \label{TTFT_model}
    \end{minipage}
    \hfill
    \begin{minipage}{0.48\linewidth}
        \includegraphics[width=\linewidth]{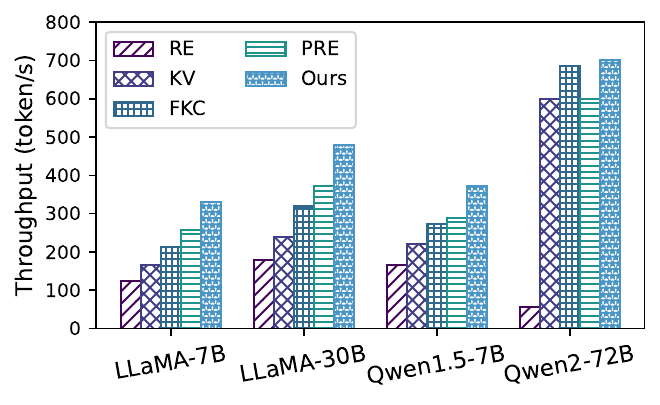}
        \caption{Prefill throughput}
        \label{throughput-model}
    \end{minipage}
\end{figure}

\noindent\textbf{Baselines.}
\begin{itemize}
    \item Full KV recompute~\cite{gao2024cost, gao2025fast}. The whole historical conversations are fed into LLM as input. The model recomputes the KV caches during the prefilling phase.
    \item Full KV load~\cite{kwon2023efficient}: The model loads the KV caches from the CPU as proposed in AttentionStore. Since we aim to test the inference system's real-time response ability, we do not use a job queue to prefetch the KV caches. Instead, we fetch the KV caches right after the request 
    arrives.
    \item Partial KV recompute~\cite{yu2025stateful, jin2024compute, yin2024llm}: The model recomputes and loads the KV caches in parallel. The recomputation ratio and loading ratio are fixed in each layer.
    \item KV eviction. We use H$_2$O~\cite{zhang2023h2o} to represent the KV caches eviction without knowing the user's future input.
    \item Fixed KV compression. MiniCache~\cite{liu2024minicache} compresses the KV caches of adjacent layers of the latter half model layers.
\end{itemize}

\subsection{Overall Performance}

The overall performance of Krul is shown in Figure \ref{overallperformance}. We can observe that although the KV caches eviction uses sparse attention based on attention weights to drop the unimportant tokens' KV caches, it still has a relatively high accuracy loss. In contrast, Krul achieves an average accuracy loss of less than 1\%. While maintaining most of the accuracy, Krul achieves an average TTFT speedup of 1.28$\times$-2.68$\times$ compared with SOTA works.\footnote{Qwen2-72B is a MQA model and the latency of recomputation is much longer, we do not show the recomputation latency in Figure \ref{overallperformance}. We show in Figure \ref{TTFT_model}.}

Krul is better than all baselines for different reasons. Compared to full KV recompute, Krul jumps over most of the prefilling phase by loading the historical conversation KV caches. Compared to KV caches eviction, although it only maintains a part of important KV caches based on the current context and has the lowest latency, the accuracy drops a lot as KV caches eviction does not taking the future user input into consideration. Therefore, KV caches eviction may destroy the information not accessed in the current context but focused on by the future input. Compared to full KV caches load, Krul is also faster in terms of restoration speed as Krul asynchronously restores the KV caches with computation and loading. Compared to partial KV recompute, Krul compressed the KV caches in the I-R layers and reduces the loading overhead. Since the speed of loading from the CPU is faster than recomputation, Krul uses the freed up time brought by the shared part KV caches to load more KV caches from the CPU, further reducing the recomputation overhead.  We give a more detailed analysis of TTFT, storage, and accuracy in the following subsections.

\begin{figure}[t]
    \setlength\belowcaptionskip{-0.5\baselineskip}
    \centering
    \begin{minipage}{0.48\linewidth}
        \includegraphics[width=\linewidth]{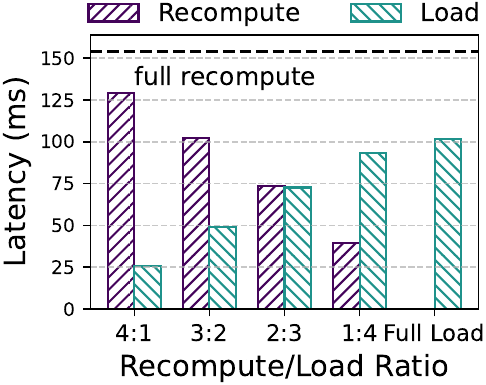}
        \caption{Latency of recomputation and loading.}
        \label{vary-ratio}
    \end{minipage}
    \hfill
    \begin{minipage}{0.48\linewidth}
        \includegraphics[width=\linewidth]{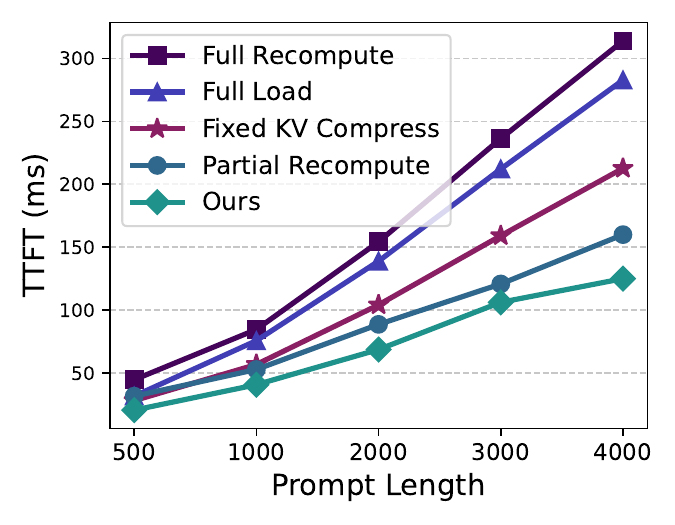}
        \caption{TTFT varying the prompt length.}
        \label{vary-length}
    \end{minipage}
\end{figure}

\subsection{TTFT and Throughput}
\label{TTFT}

\begin{table*}[t]
\caption{Evaluation results of Minicache and Krul on LongBench. }
\label{detail-evaluation}
{\setlength{\tabcolsep}{3pt} 
\begin{tabular}{@{}ccccccccc@{}}
\toprule
             & narrativeqa & multifieldqa\_zh & passretriev\_zh & 2wikimqa & qasper & multifieldqa\_en & trec & dureader \\ \midrule
Baseline & 18.96       & 46.41            & 10.0  & 12.51    & 19.7   & 40.19            & 65.5 & 34.54    \\
Minicache    & 18.49       & 45.92            & 6.12  & \textbf{13.74}    & 16.83  & 41.12            & \textbf{65.0} & 33.89    \\
Krul (Fix) & 17.67 & 44.83 & \textbf{8.5} & 11.32 & \textbf{17.1} & 40.22 & 62.5 & 33.87 \\
Krul         & \textbf{18.65}       & \textbf{45.97}            & \textbf{9.5}  & 12.54    & \textbf{18.46}  & \textbf{43.24 }           & 64.0 & \textbf{34.16}    \\ \bottomrule
\end{tabular}
\\[0.5cm]  

}
\begin{tabular}{ccccccccc}
\toprule
             & gov\_report & vcsum & lsht & qmsum  & multi\_news & passretriev\_en & samsum & Average. \\ \midrule
Baseline & 25.3        & 16.44 & 19.5 &    21.36                & 25.92       & 8.0                    & 39.78 & 26.94  \\
Minicache    & 20.06       & 15.96 & 20.0 &       21.32             & 23.24       & 8.0                    & 37.62 & 25.82  \\
Krul (Fix) & \textbf{21.59} & \textbf{15.97} & 18.5 & 21.05 & \textbf{23.79} & 8.0 & \textbf{40.9} & 25.72 \\
Krul        & \textbf{23.24}       & \textbf{16.21} & \textbf{20.5} &      \textbf{21.51}              & \textbf{24.26 }      & 8.0                    & \textbf{40.4} & \textbf{26.70}   \\ \bottomrule
\end{tabular}
\end{table*}

TTFT is an important metric for the quality of service in LLM inference. It indicates how quickly users start seeing the output of LLMs after entering their prompt. As shown in Figure \ref{TTFT_model}, Krul can reduce the TTFT by an average of  $2.65\times$ in LLaMA-7B, 30B, and Qwen1.5-7B and $6.62\times$ in Qwen2-72B compared with full KV recomputation (RE). Krul achieves an average of $2.35\times, 1.76\times$ speedup compared with full KV loading (KV), fixed KV compression (FKC). Additionally, Krul achieves a $1.5\times$ speedup compared with partial KV recompute (PRE). This is because Krul compresses the KV caches and reduces both the recomputation and loading latency.

Prefilling throughput is the metric to evaluate the speed of processing the prompt. We observe that Krul provides an average throughput speedup of $2.5\times$, $1.9\times$, $1.41\times$, and $1.29\times$ compared with full recompute, full KV load, fixed KV compression, and partial recompute. The improvement of Krul comes from the parallel KV caches restoration with recomputation and loading. Moreover, Krul compresses the KV caches and rearranges the restoration pipeline to further improve the prefilling speed.

We provide an in-depth analysis of varying  KV caches recompute/load ratio and the historical conversation length on LLaMA-7B. Figure \ref{vary-ratio} shows the recomputation and KV caches loading latencies with different recompute/load ratios. The TTFT depends on the longer of them. The computation latency decreases and the loading latency increases with a higher ratio of KV caches loading. The ratio 2:3 achieves an almost recomputation-loading overlap and provides $2\times$ and $1.35\times$ speedup compared with full KV recompute and load. Figure \ref{vary-length} shows the TTFTs varying the prompt length. Krul has a slower TTFT increasing speed compared with the other three methods. This improvement comes from the KV caches compression and the recomputation-loading pipeline of KV caches restoration.

\begin{figure}[t]
    \vspace{-8pt}
    \setlength\belowcaptionskip{-0.5\baselineskip}
    \centering
    \begin{minipage}{0.51\linewidth}
        \includegraphics[width=\linewidth]{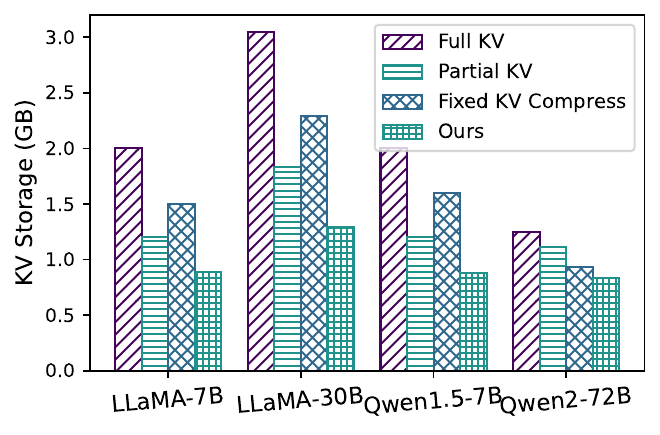}
        \caption{KV caches storage varying different models.}
        \label{KVStorage}
    \end{minipage}
    \hfill
    \begin{minipage}{0.45\linewidth}
        \includegraphics[width=\linewidth]{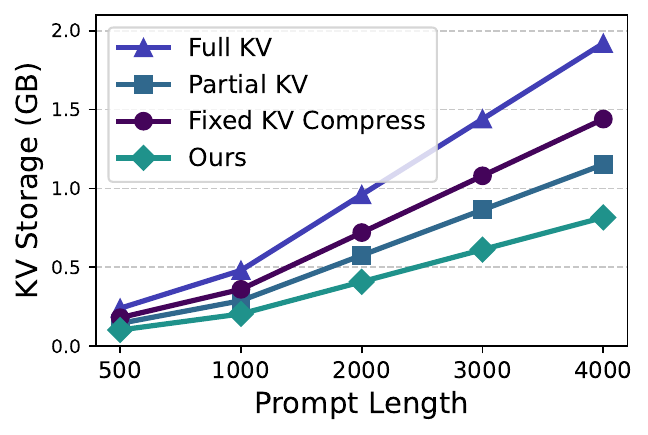}
        \caption{KV caches storage varying different prompt lengths.}
        \label{KVStoragevary-length}
    \end{minipage}
    \vspace{-8pt}
\end{figure}

\subsection{KV Caches Storage}

KV caches storage is the metric to evaluate the memory usage when the conversation is inactive. The result is shown in Figure \ref{KVStorage}. Our Krul achieves an average of 2.35$\times$ and $1.75\times$ KV caches reduction compared with full KV storage and fixed KV compression. The improvement comes from two aspects: 1) Krul recomputes a part of the historical conversation KV caches, so there is no need to store the whole KV caches. 2) Krul compressed the KV caches in I-R layers, enabling multiple tokens to use the same KV caches to store the information of historical conversations. 

We provide an in-depth analysis of varying the historical conversation length on LLaMA-7B.  Figure \ref{KVStoragevary-length} shows the KV caches memory space varying the prompt length. We can observe that Krul has a slower KV caches storage increasing speed compared with the other two methods, which can effectively compress KV caches space when restoring the KV caches of long conversations. This improvement comes from the recomputation-loading-mixed KV restoration and KV caches compression of KV caches restoration.

\subsection{Accuracy}

\label{Accuracy}

We provide a detailed accuracy result of Qwen1.5-7B in Table \ref{detail-evaluation}. Minicache fixed the shared layer range to the latter half layers of the model, which are part of the I-R layers. And Minicache limits the share strategy to adjustment layers.  Therefore, Minicache's compression strategy is a subset of our Krul. Our KV caches compression formula is the same as that of Minicache. However, we scale the shared layer range to all I-R layers and calculate dynamic compression strategies for different historical conversations. Therefore, we achieve a high accuracy compared with Minicache. In some tasks, such as passretiev\_zh, Minicache has a relatively high accuracy loss, while our Krul still maintains most of the accuracy.

\begin{figure}[t]
    \setlength\abovecaptionskip{0.02\baselineskip}
    \setlength\belowcaptionskip{-0.5\baselineskip}
    \includegraphics[width=0.6\linewidth]{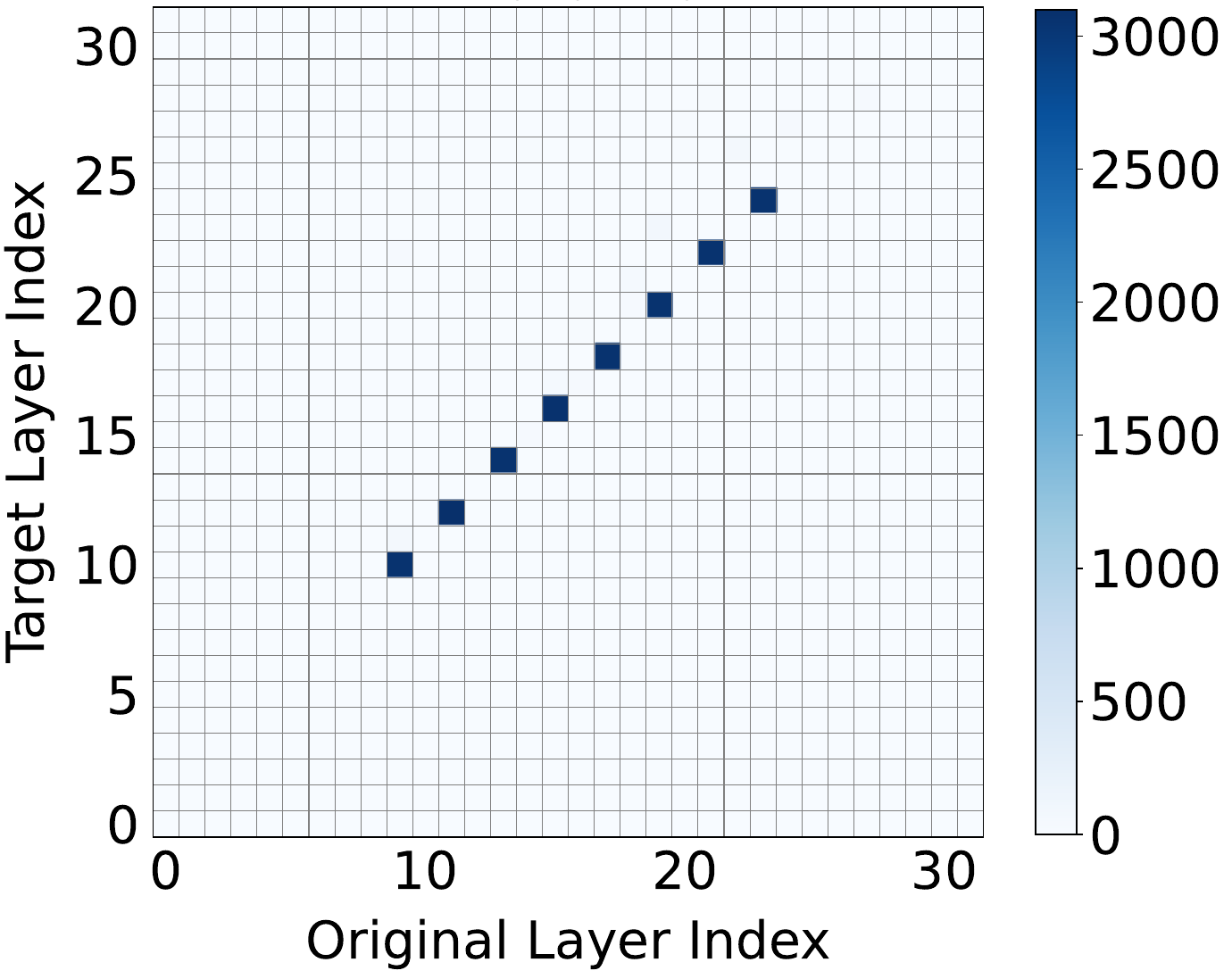}
    \caption{The compression map of different historical conversations.}
    \label{sharemap}
\end{figure}

\textbf{In-depth analysis.} We analyze the strategies constructed from Krul. We fix the total shared layer number to half of the total layers, which is the same as Minicache. We use 3K data from ShareGPT and count the shared layer pairs. The result is shown in Figure \ref{sharemap}. Although we scale the searching space to all I-R layers, most strategies still share the adjustment layers. That means the attention weights between the adjustment layers are usually the most similar. However, the biggest difference from Minicache is that the most shared layer range is layer 9-24, while in Minicache is the latter half of the layers (layer 16-31).

Based on this observation, we fix the shared layer to every two adjustment layers in 9-24 and supplement the result in Table \ref{detail-evaluation}. We can observe that even if the shared layers are in the former half of the model, it still maintains most of the accuracy as long as the shared layers are I-R, which is different from the conclusion in Minicache. It achieves less accuracy loss than Minicache in some tasks. In contrast, Krul provides dynamic compression strategies for different historical conversations to minimize the accuracy loss. From this point, we can also confirm that using attention weight to measure the similarities between the model layers is a good choice while using KV caches compression.

\subsection{Extra Overhead Analysis}

\begin{figure}[t]
    \setlength\belowcaptionskip{-0.5\baselineskip}
    \vspace{-8pt}
    \includegraphics[width=0.7\linewidth]{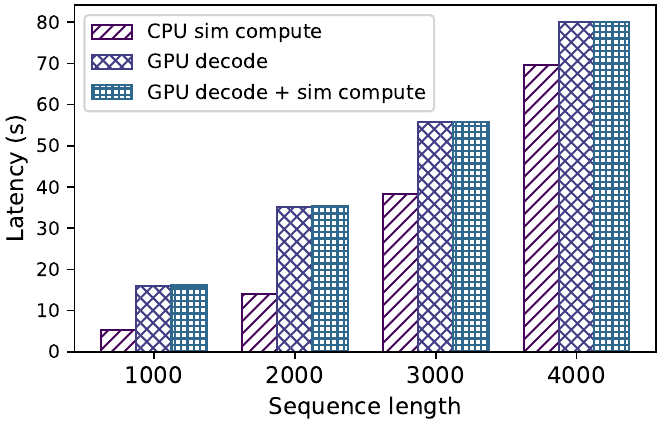}
    \caption{The comparison of latency of CPU attention similarity computing, GPU decoding and GPU decoding with similarity computing.}
    \label{overhead_analysis}
\end{figure}

We analyze the extra overhead brought by KV caches compression. We asynchronously compute the similarity of attention weights of the prefilling phase in the CPU while decoding and computing the similarity of the decoding phase in the GPU. The latency is shown in Figure \ref{overhead_analysis}. We can observe that the latency of CPU computing is less than GPU decoding when processing a sequence of the same length. Moreover, as we analyze in Section \ref{Heterogeneous},  the decoding length has an average of 6.56$\times$ longer than the prefilling length, which ensures the similarity result computed in CPU can be obtained before the end of decoding. 

In addition, since we compute the similarity of the decoding phase in the GPU, it also brings extra computation overhead in the decoding phase. However, as shown in Figure \ref{overhead_analysis}, the computation overhead brought by computing similarity in the GPU can almost be ignored, since we only reserve the attention weight of the last token during the decoding phase. In summary, the computation and storage overheads are overlapped by our heterogeneous similarity computing.

\section{Related Work}

\textbf{State Restoration}. The model needs to restore the conversation state by getting all the KV caches of the historical tokens. At the beginning, vLLM~\cite{kwon2023efficient} and DeepSpeed~\cite{rasley2020deepspeed} recompute all the historical tokens. As a result, the computation cost increases sharply as the sequence length increases. Then vLLM and AttentionStore~\cite{gao2024cost} store the KV caches across conversation turns, which we call the prefix cache. The prefix cache compares the requests at the front of the request queue with the current processing requests, then loads the different parts of the KV caches from the CPU into the GPU asynchronously. These approaches jump over most prefill stage.  Hcache~\cite{gao2025fast} reduces the KV caches loading cost by storing the hidden state across different conversation turns and recomputing the KV caches using the hidden state. Hcache always reduces half of the KV caches storage space. Since the MQA model already reduces more than $4\times$ KV caches storage through the grouping of KV heads, making HCache adjust to the MQA models requires the modification of the hidden state representation. Pensieve~\cite{yu2025stateful}, Cake~\cite{jin2024compute}, and MobileLLM~\cite{yin2024llm} propose a suffix cache compute-loading pipeline. They recompute the front part and load the back part of the KV caches asynchronously. This approach reduces the loading cost and jumps over part of the prefilling stage, which is a compromise between full recomputation and prefix cache. Different from all the approaches above, Krul identifies an opportunity to compress the KV caches across conversation turns. Krul not only reduces the loading cost but also reduces the KV caches memory cost after the prefilling stage, which means Krul enables a larger batch size compared with the approaches above. Moreover, our KV caches compression method can directly adjust to MQA models.

\textbf{Sparse Attention}. Existing works use sparse attention to reduce the KV caches during the decoding phase. StreamingLLM ~\cite{xiao2023efficient} just keeps the recent tokens' KV caches. $\text{H}_2\text{O}$~\cite{zhang2023h2o}, Q-hitter~\cite{zhang2024q} uses a heavy hitter to dynamically filter the important token based on the attention weight and evicts other KV caches. Infinigen~\cite{lee2024infinigen}, Quest~\cite{tang2024quest} store all of the KV caches and dynamically prefetch the target KV caches based on the current context. Although sparse attention is memory-efficient, it cannot be directly used in the multi-turn conversation since it cannot take the possible user's future input into consideration, which may compromise the generation quality of the model. In contrast, instead of evicting or saving the entire KV caches, Krul saves the back part of the KV caches and compresses them in the I-R layers, which reduces the KV caches while retaining all key information contained in the KV caches. Therefore, Krul reserves most of the accuracy of the model.

\textbf{KV Caches Compression}. Existing works inter-layer and intra-layer KV caches compression to reduce the KV caches memory cost during the decoding phase. KVMerger~\cite{wang2024model} and CAM~\cite{zhang2024cam} merge KV caches with high cosine similarity in the same layer and make multiple tokens share the same KV caches. Minicache~\cite{liu2024minicache} merges the KV caches in the deep layers. Minicache's compression strategy space is limited to every two adjacent layers, which may miss out on better compression methods. KVSharer calculates the Euclidean distance between the KV caches of every possible layer combination. It then arranges these layer pairs according to their degree of dissimilarity, giving precedence to compression among the most dissimilar layers. However, this approach needs a sample dataset and performs multiple forward processes to find a fixed share policy for all the data. As a result, the characteristics of different data may be ignored. In contrast, Krul customizes different compression and compression strategies for different datasets from a broader strategy space, which leads to a smaller loss of accuracy. 

\section{Conclusion}

In this paper, we present Krul, a novel LLM inference system with a conversation-adaptive KV cache compression mechanism. To improve the restoration efficiency while not compromising the generation quality and performance, we design a preemptive compression strategy
selector, a token-wise
heterogeneous attention similarity estimator, and a bubble-free restoration scheduler to address the accuracy, storage, and computation problems brought by KV caches compression in the multi-turn conversation scenario. Extensive experimental results demonstrate that Krul decreases the TTFT by an average of $1.28\times\sim 2.68\times$ and storage space by $1.33\times \sim 2.35\times$ compared with the state-of-the-art state restoration works without compromising the generation quality.



\bibliographystyle{ACM-Reference-Format}
\bibliography{sample}

\end{document}